\documentclass[twoside]{article}

\usepackage[accepted]{aistats2026}
\usepackage[round]{natbib}

\usepackage[T1]{fontenc}
\usepackage[utf8]{inputenc}
\usepackage{microtype}

\usepackage{amsmath,amssymb,amsthm,mathtools,bbm}

\usepackage{graphicx}
\usepackage{booktabs}
\usepackage{subcaption} 
\usepackage{tabularx}
\usepackage{adjustbox}
\usepackage{multirow}

\usepackage[dvipsnames]{xcolor}
\usepackage{todonotes}

\usepackage{hyperref}
\hypersetup{colorlinks=true,linkcolor=MidnightBlue,citecolor=MidnightBlue,urlcolor=MidnightBlue}

\usepackage{algorithm}
\usepackage[noend]{algpseudocode}

\newtheorem{thm}{Theorem}[section]
\newtheorem{lem}[thm]{Lemma}

\newtheorem{exmp}[thm]{Example}
\newtheorem{prop}[thm]{Proposition}
\newtheorem{cor}[thm]{Corollary}
\newtheorem{assum}[thm]{Assumption}
\theoremstyle{definition}

\theoremstyle{remark}
\newtheorem{rem}[thm]{Remark}
\theoremstyle{condition}

\DeclareMathOperator{\E}{\mathbb{E}}
\DeclareMathOperator{\Var}{Var}
\DeclareMathOperator{\Cov}{Cov}
\DeclareMathOperator{\Tr}{Tr}

\newcommand{\cH}{\mathcal{H}}

\newcommand{\cI}{\mathcal{I}}
\newcommand{\cX}{\mathcal{X}}
\newcommand{\cA}{\mathcal{A}}
\newcommand{\cY}{\mathcal{Y}}
\newcommand{\cF}{\mathcal{F}} 

\newcommand{\cD}{\mathcal{D}}

\newcommand*\diff{\mathop{}\!\mathrm{d}}

\usepackage{enumitem}

\usepackage{enumitem}

\begin{document}

%

%

\twocolumn[

\aistatstitle{Kernel Treatment Effects with Adaptively Collected Data}

\aistatsauthor{ Houssam Zenati$^{1}$ \And Bariscan Bozkurt$^{1}$ \And Arthur Gretton$^{1,2}$ }
\vspace{0.5cm}
\aistatsaddress{ \hspace{4.2cm}$^{1}$Gatsby Computational Neuroscience Unit, University College London \And \hspace{4cm}$^{2}$Google DeepMind } 

\runningauthor{Houssam Zenati, Bariscan Bozkurt, Arthur Gretton}
]

\begin{abstract}
Adaptive experiments improve efficiency by adjusting treatment assignments based on past outcomes, but this adaptivity breaks the i.i.d.\ assumptions that underpin classical asymptotics. At the same time, many questions of interest are distributional, extending beyond average effects. Kernel treatment effects (KTE) provide a flexible framework by representing interventional outcome distributions in an RKHS and comparing them via kernel distances. We present the first kernel-based framework for distributional inference under adaptive data collection. Our method combines doubly robust RKHS scores with a witness function learned on one fold, and performs inference on a second fold using a projected, sequentially normalized scalar statistic with valid type-I error.  Experiments show that the resulting procedure is well calibrated and effective for both mean shifts and higher-moment differences, outperforming adaptive baselines limited to scalar effects.
\end{abstract}

\section{INTRODUCTION}

Data in modern experiments are increasingly collected \emph{adaptively}, with treatment assignments chosen sequentially in response to past outcomes, as in multi-armed and contextual bandits \citep{LattimoreSzepesvari2020}, adaptive clinical trials \citep{ChowChang2011}, and dynamic pricing strategies in economics \citep{QiangBayati2016,AtheyEtAl2022}. Adaptivity improves participant welfare and accelerates learning during data collection, but it fundamentally alters the statistical properties of the data: allocation proportions and effective sample sizes become random and history-dependent \citep{bibaut2025demystifying}. This breaks the classical i.i.d.\ assumptions that underlie standard asymptotic theory \citep{vaart1997weak}, and as a consequence, estimators that are asymptotically normal under fixed designs may converge to non-normal limits or exhibit inflated variances \citep{bibaut2025demystifying}.

Simultaneously, reliance on the average effects is often insufficient, as many scientific and practical questions are inherently \emph{distributional}. In medicine, clinicians care not only about mean efficacy but also about the distribution of side effects across patients \citep{rothe2010nonparametric}; in finance and operations, decision-makers evaluate policies using tail-sensitive criteria such as conditional value-at-risk (CVaR) \citep{rockafellar2000optimization}; and in reinforcement learning, distributional approaches explicitly target higher moments or quantiles of return distributions \citep{dabney2018implicit}. Existing statistical methods often rely on cumulative distribution functions \citep{chernozhukov2013inference,huang2021}, which become difficult to extend to high-dimensional or structured outcomes.\looseness=-1

Kernel methods provide a powerful alternative. Kernel mean embeddings can represent interventional outcome distributions as elements of a reproducing kernel Hilbert space (RKHS) \citep{berlinet2011reproducing,gretton2013introduction, muandet2021counterfactual}, enabling nonparametric comparison of distributions via kernel distances and supporting inference on complex outcomes such as images, sequences, or graphs \citep{gartner2003survey}. This framework has been used to define conditional distributional kernel treatment effects \citep{park2021conditional}, to design kernel-based hypothesis tests \citep{shekhar2023, martinez2023,fawkes2024doubly}, and to extend efficiency theory to Hilbert-space parameters \citep{luedtke2024}. However, all existing KTE methods assume i.i.d.\ data, and it remains unknown how to conduct distributional causal inference when outcomes are observed under adaptive, history-dependent policies.

In this paper, we develop the first framework for \emph{kernel treatment effect inference under adaptive data collection}. Our contributions are as follows: \textit{i)} we construct a doubly robust procedure that learns an RKHS witness on a first chronological fold and performs inference on a second fold using projected kernel scores and sequential normalization based only on past data; \textit{ii)} we develop a reweighted plug-in estimator of the projected conditional standard deviation and prove its consistency under adaptive logging; \textit{iii)} we obtain a sample-split adaptive test with valid Gaussian limits under the null; and \textit{iv)} we validate the method in numerical simulations, including semi-synthetic benchmarks and high-dimensional structured outcomes such as images, showing that it detects distributional effects missed by scalar adaptive baselines.

Conceptually, our work connects kernel-based distributional causal inference with adaptive inference. Rather than deriving a Gaussian limit for the full RKHS-valued effect\footnote{An earlier version of this work pursued a Gaussian limit for the difference of interventional mean embeddings themselves, rather than for the test statistic directly. See Appendix \ref{app:why_not_rkhs_stabilization}.}, we reduce inference to a scalar adaptive statistic while retaining sensitivity to distributional alternatives through a learned witness.

The remainder of the paper is structured as follows. Section \ref{sec:related_works} reviews related work. Section \ref{section:kte_adaptive} formalizes the adaptive setting and KTE. Section \ref{section:adaptive_test} introduces our adaptive test statistic. We detail plug-in variance estimation in Section \ref{section:variance-estimation} and the practical procedure and power guarantees in Section \ref{section:practical_test_power}. Section \ref{section:experiments} reports simulations, and Section \ref{section:discussion} concludes.\looseness=-1

\section{RELATED WORKS}
\label{sec:related_works}

Kernel mean embeddings \citep{smola2007hilbert} provide a nonparametric representation of distributions in RKHS and a way to compare them through inner products and norms \citep{kanagawa14,SriGreFukLanetal10,gretton12a}. Building on this framework, \citet{muandet2021counterfactual} introduced \emph{Counterfactual Mean Embeddings} to represent full interventional outcome distributions and define distributional treatment effects under unconfoundedness; related work expressed ATE and CATE through conditional mean embeddings, yielding RKHS formulations of heterogeneous effects \citep{park2021conditional,singh2024kernel}. On the inferential side, \citet{fawkes2024doubly} proposed doubly robust kernel statistics for testing equality of interventional outcome distributions, \citet{martinez2023} developed an efficient doubly robust kernel test with improved power and valid type-I error control, and \citep{luedtke2024,zenati2025cpme} established estimation guarantees for doubly robust estimators of interventional mean embeddings in non-adaptive settings. 

Adaptive experimentation arises in multi-armed bandits, best-arm identification, adaptive clinical trials, contextual bandits, batch bandits, sequential policy learning, and dynamic pricing \citep{LattimoreSzepesvari2020,GarivierKaufmann2016,ChowChang2011, zenati22a, LiChuLangfordSchapire2010,perchet2016,zenati2023sequential,QiangBayati2016}. While such designs improve performance during data collection, they make inference harder because allocation proportions and effective sample sizes become random and history-dependent \citep{AtheyEtAl2022,CariaEtAl2023}, breaking the classical i.i.d.\ asymptotic framework \citep{vanderVaart1998asymptotic,hall1980}. For adaptive inference, \citet{hadad2021confidence} showed that suitable reweighting can recover approximate normality for policy evaluation, while related stabilization approaches include conditional-variance weighting and adaptive weighting without outcome models \citep{bibaut2021post,zhang2021statistical}; see also \citet{zhang2020inference} for batched bandits, \citet{howard2021time_uniform,waudby_smith2021tu_clt} for always-valid inference, and \citet{bibaut2025demystifying,hirano2023asymptotic} for a broader view of when Gaussian limits fail or can be restored. Our work extends this literature to \emph{distributional kernel treatment effects} under contextual adaptivity by combining a sample-split kernel witness, projected doubly robust scores, and sequential normalization.

\section{PROBLEM STATEMENT}
\label{section:kte_adaptive}

We formalize the estimation of \emph{kernel treatment effects} (KTE) when data are collected via an \emph{adaptive experiment} (e.g., contextual bandit algorithm). \looseness=-1

\subsection{Adaptive data collection setting}

We consider a contextual decision-making system operating over \( T \) rounds. At each round \( t \in \{1, \dots, T\} \), the agent observes a context \( X_t \in \cX \), sampled independently from an unknown distribution \( P_X \), i.e., \( X_t \sim P_X \). Given \( X_t \), the agent selects an action \( A_t \in \cA \) according to a possibly adaptive policy \( \pi_t \in \Pi \), such that \( A_t \sim \pi_t(\cdot \mid \cF_{t-1}, X_t) \), where \( \cF_{t-1} := \sigma(X_1, A_1, Y_1, \ldots, X_{t-1}, A_{t-1}, Y_{t-1}) \) denotes the filtration up to time \( t-1 \). The outcome \( Y_t \in \cY \) is then generated according to a fixed, unknown outcome model \( Y_t \sim P_{Y \mid X, A}(\cdot \mid X_t, A_t) \), depending only on the current context and action. We assume that the action space \( \cA \) is discrete and the outcome space \( \cY \) may be either discrete or continuous, and that each policy \( \pi_t \) admits a density with respect to a base measure \( \mu_{\cA} \). The sequence of policies \( \{\pi_t\}_{t=1}^T \) may depend on past observations, rendering the overall data-generating process adaptive rather than i.i.d. The observed dataset consists of the trajectory \( \mathcal{D}_T = \{(X_t, A_t, Y_t)\}_{t=1}^T \). We assume the existence of a potential outcome function \( a \mapsto Y_t(a) \) such that \( Y_t = Y_t(A_t) \), and that the collection \( \{Y_t(a)\}_{a \in \cA} \) is conditionally independent of \( A_t \) given \( X_t \), i.e., conditional ignorability holds. \looseness=-1

\subsection{Target Estimand}

We work throughout with an outcome kernel \(k_\cY\) satisfying the following standing condition.

\begin{assum}[Outcome kernel]
\label{assumption:bounded_kernel}
The kernel \(k_\cY\) is bounded and characteristic. That is, there exists \(\kappa<\infty\) such that $\sup_{y\in\cY} k_\cY(y,y)\le \kappa$,
and the associated mean embedding of probability measures into \(\cH_\cY\) is injective.
\end{assum}
Let \( \cH_\cY \) denote the RKHS associated with \( k_\cY \), and let \( \phi_\cY(y) = k_\cY(\cdot, y) \) be its feature map. We first introduce the \textit{interventional mean embedding} (IME) \citep{muandet2021counterfactual}\footnote{We prefer the term \emph{interventional mean embedding} because our estimand is a population-level intervention distribution \(P(Y \mid do(A=a))\), not a counterfactual quantity in Pearl’s stricter rung-3 sense.} of the interventional outcome distribution of \(Y(a)\) for \(a \in \cA\):
\begin{equation}
    \eta(a):= \mathbb{E}_{ P_X \times P_{Y \mid X, A}} \left[ \phi_\cY(Y(a)) \right], \tag{IME}
    \label{eq:IME}
\end{equation}
where the expectation is taken over the fixed marginal distribution of contexts $P_X$ and the conditional outcome distribution $P_{Y|X,A}$. Then, the \textit{target estimand} is the generalized \textit{kernel treatment effect} \eqref{kte:psi_definition} which is defined as expressed as the MMD of the two interventional mean embeddings \( \eta(a) \) and \( \eta(a') \), that is the RKHS norm of the difference $\Psi$:
\begin{align}
    \tau(a,a')&:= \Vert \Psi(a,a') \Vert_{\cH_\cY}, \label{kte:psi_definition} \tag{KTE}\\ 
    \Psi(a,a')&:=\eta(a)-\eta(a'), \label{eq:psi_definition} 
\end{align}
Because \(k_\cY\) is characteristic, \(\tau(a,a')=0\) if and only if the two interventional outcome distributions under \(a\) and \(a'\) coincide. Now, define the following conditional mean embedding  (CME) \citep{song2009hilbert} of the distribution \( P_{Y\mid X, A} \):
\begin{equation}
\mu_{Y\mid A,X}(a, x):=\mathbb{E}_{P_{Y\mid X, A}}[\phi_{\cY}(Y) \mid A=a, X=x].
\label{eq:conditional_mean_embedding}
\end{equation}
\begin{assum}[Sequential ignorability and positivity]
\label{assum:selection_observables}
Fix the target actions $a,a'\in\cA$. For each $b\in\{a,a'\}$ and each $t\ge 1$: i) \textit{Consistency:} $Y_t=Y_t(b)$ on the event $\{A_t=b\}$. ii) \textit{Sequential exchangeability:} $Y_t(b)\perp A_t\mid (X_t,\cF_{t-1})$. iii) \textit{Positivity:} $\pi_t(b\mid X_t)>0$ almost surely.
\end{assum}
Under Assumption \ref{assum:selection_observables}, the \eqref{eq:IME} can be identified from observable data  \citep{muandet2021counterfactual, zenati2025cpme}:
\begin{equation}
\eta(a)= \mathbb{E}_{P_X}\left[\mu_{Y\mid A,X}(a, x)\right].
\label{eq:identified_cpme}
\end{equation}
To construct doubly robust scores, let $\pi$ be any conditional density on $\cA\times\cX$ and let $\bar\mu:\cA\times\cX\to\cH_{\cY}$ be any measurable function. For discrete $\cA$, define the uncentered doubly robust (DR) score
\begin{equation}\label{eq:canonical_gradient_revised}
\begin{aligned}
D'(\pi,\bar\mu,a)(X,A,Y)
:=
&\frac{\mathbbm 1\{A=a\}}{\pi(a\mid X)}\Bigl(\phi_{\cY}(Y)-\bar\mu(a,X)\Bigr) \\
&+
\bar\mu(a,X).
\end{aligned} 
\end{equation}
Its centered version
\[
D(\pi,\bar\mu,a):=D'(\pi,\bar\mu,a)-\eta(a)
\]
coincides with the canonical gradient of the \eqref{eq:IME} \citep{luedtke2024}. In what follows, we work directly with the uncentered score $D'$, since its conditional expectation equals the target embedding.
\subsection{Why naive kernel tests fail under adaptivity}
\label{section:failure_standard_asymptotic}

Let \( ( \widehat{\mu}_{Y \mid A,X}^{(t)} )_{t \ge 1} \) denote a sequence of estimators of the conditional mean embedding \( \mu_{Y \mid A,X} \), where each \( \widehat{\mu}_{Y \mid A,X}^{(t)} : \cA \times \cX \to \cH_{\cY} \) is trained using data up to round \( t \), and define the doubly robust RKHS score difference
\begin{equation}
\label{eq:phi_t_def}
\begin{aligned}
\hat{\phi}_t
:= {} & D^{\prime}(\pi_t,\widehat{\mu}_{Y \mid A,X}^{(t-1)},a)(X_t,A_t,Y_t) \\
&- D^{\prime}(\pi_t,\widehat{\mu}_{Y \mid A,X}^{(t-1)},a')(X_t,A_t,Y_t)
\in \cH_{\cY}.
\end{aligned}
\end{equation}
In i.i.d.\ settings, averages of \((\hat{\phi}_t)\) admit Gaussian fluctuations in \(\cH_{\cY}\) under standard conditions. However, for testing
\[
H_0:\Psi(a,a')=0,
\]
the relevant object is a quadratic functional of this average, and the direct plug-in statistic
\[
\big\|\widehat{\Psi}_T(a,a')\big\|_{\cH_{\cY}}^{2}
\]
is degenerate under the null, as in standard MMD testing. Accordingly, even in the i.i.d.\ case, valid kernel tests rely on sample splitting and studentization of a cross-\(U\) statistic \citep{kim2024dimension}. Under adaptive data collection, there is an additional difficulty: because \(\pi_t\) depends on the past, the sequence \((\hat{\phi}_t)\) is no longer i.i.d., and its conditional covariance
\[
\Sigma_t := \Cov(\hat{\phi}_t \mid \cF_{t-1})
\]
varies with time. Thus the usual i.i.d.\ calibration need not survive under adaptive logging, even when the propensities are known. A scalar normalization based only on \(\Tr(\Sigma_t)^{-1/2}\) controls the overall scale of the covariance operator but not its directional variation in \(\cH_{\cY}\). Since our goal is valid inference for a final scalar test statistic rather than a full Gaussian limit in \(\cH_{\cY}\) (see Appendix \ref{app:why_not_rkhs_stabilization}), this motivates projecting onto a learned witness and applying a sequential, past-measurable variance normalization to the resulting scalar score process.

\begin{figure}[t]
  \centering
  \includegraphics[width=0.7\columnwidth]{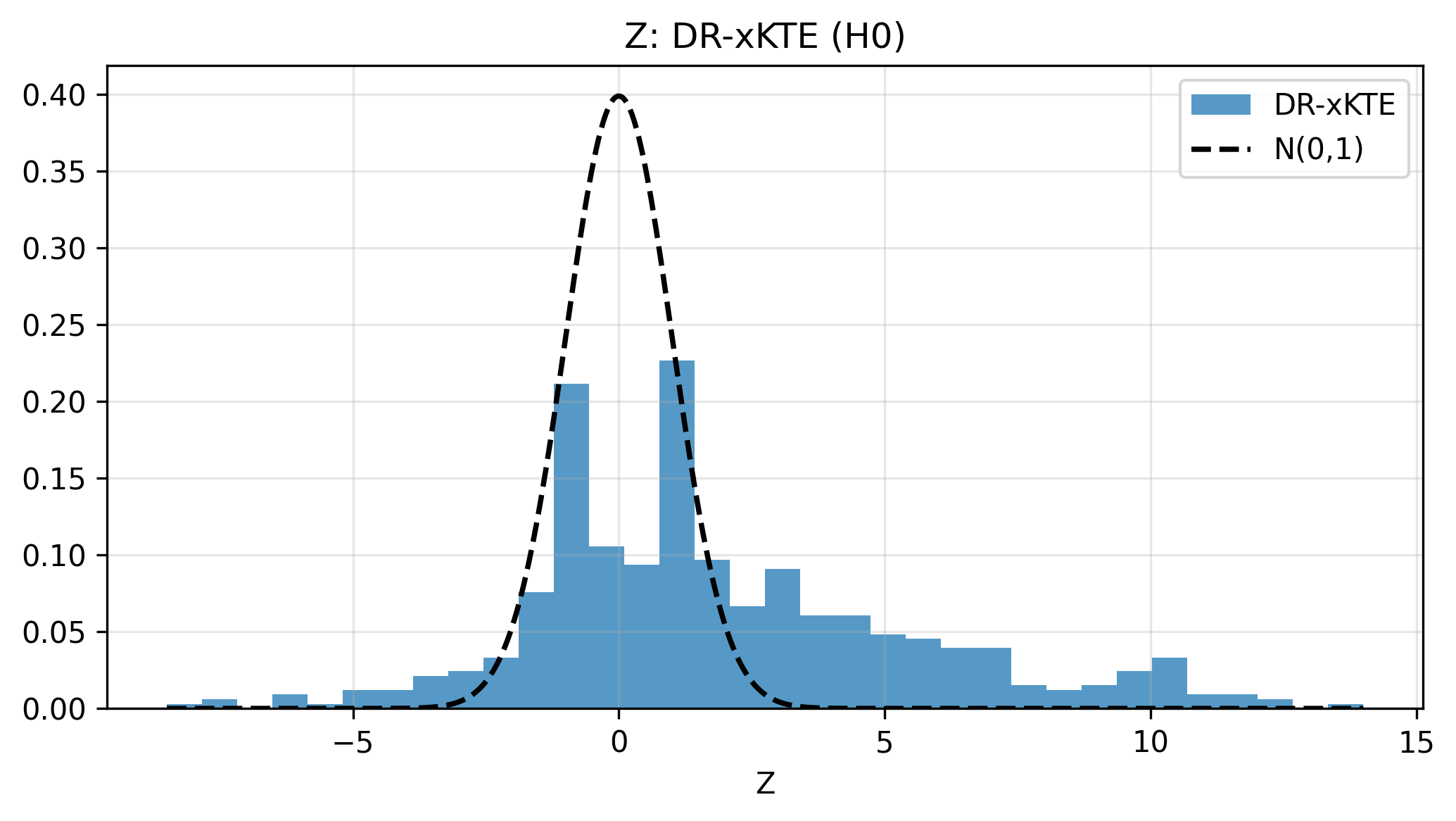}
  \caption{Histogram of the i.i.d.-style DR-xKTE statistic over \(500\) runs
  (\(T=700\), \(d=5\), \(t_0=15\), \(\varepsilon=10^{-3}\)) under adaptive logging with known propensities \(\pi_t(1 \mid X_t)\). The resulting null distribution is visibly miscalibrated.}
  \label{fig:etc-failure}
\end{figure}

\begin{exmp}[Contextual extension of \citep{bibaut2025demystifying}, Section~3.1.1]
\label{exmp:etc}
Let \(X_t \stackrel{\text{i.i.d.}}{\sim} \mathcal{N}(0,I_d)\), and let
\[
Y_t(0)=f(X_t)+\varepsilon_t,
\qquad
Y_t(1)=f(X_t)+\Delta(X_t)+\varepsilon_t,
\]
where \(\Delta=Y_t(1)-Y_t(0)\) is the treatment shift. The policy explores uniformly for \(t \le t_0\), then commits to the empirically better arm with \(\varepsilon\)-randomization:
\[
\pi_t(1\mid X_t)=
\begin{cases}
0.5, & t\le t_0,\\
1-\varepsilon, & t>t_0 \text{ and arm }1\text{ is selected},\\
\varepsilon, & t>t_0 \text{ and arm }0\text{ is selected}.
\end{cases}
\]
Since the committed arm is history-dependent, the design is adaptive. Under \(H_0:\Delta\equiv 0\), \citet{bibaut2025demystifying} show that even the scalar ATE can have a non-Gaussian mixture limit. Figure~\ref{fig:etc-failure} shows that the naive i.i.d.-style DR-xKTE statistic of \citet{martinez2023} is similarly miscalibrated in this adaptive regime.
\end{exmp}

Example~\ref{exmp:etc} and Figure~\ref{fig:etc-failure} illustrate the issue addressed in this paper: under adaptive treatment assignment, the usual i.i.d.\ studentization of a kernel test is no longer sufficient, and valid inference must account for the time-varying conditional variability of the score sequence using only past information.

\section{ADAPTIVE TEST STATISTIC}
\label{section:adaptive_test}

We now introduce the adaptive test statistic studied in this paper. The construction has two stages. A first chronological fold is used to learn a witness in \(\cH_{\cY}\) for the direction of \(\Psi(a,a')\). A second fold is then used for inference after projecting the doubly robust RKHS score onto that witness and normalizing the resulting scalar process by a predictable estimate of its conditional standard deviation.

\subsection{Pilot witness and projected score}

Throughout this section, write \(T=2n\) and use the chronological split
\[
\mathcal I_1:=\{1,\dots,n\},
\qquad
\mathcal I_2:=\{n+1,\dots,2n\}.
\]
The first fold is used to construct the witness, while inference is carried out on the second fold.

For \(r\in\{1,2\}\) and \(t\in \mathcal I_r\), let \(\widehat\mu^{(r)}_{t-1}\) be an \(\cF_{t-1}\)-measurable estimator of \(\mu_{Y\mid A,X}\), and define the foldwise doubly robust RKHS score
\begin{equation}
\label{eq:phi_t_fold}
\begin{aligned}
\hat\phi_t^{(r)}
:= {} &
D'(\pi_t,\widehat\mu^{(r)}_{t-1},a)(X_t,A_t,Y_t) \\
&-
D'(\pi_t,\widehat\mu^{(r)}_{t-1},a')(X_t,A_t,Y_t)
\in \cH_{\cY}.
\end{aligned}
\end{equation}

Throughout the analysis, we impose the standing envelope (e.g when learning a conditional mean embedding with fixed regularization parameter \citep{li2022optimal})
\begin{equation}
\label{eq:nuisance_envelope}
\sup_{r\in\{1,2\}}
\sup_{t\ge1}
\sup_{b\in\cA,\ x\in\cX}
\big\|
\widehat\mu^{(r)}_{t-1}(b,x)
\big\|_{\cH_{\cY}}
\le B_\mu
\quad \text{a.s.}
\end{equation}

The pilot witness is the first-fold average
\begin{equation}
\label{eq:pilot_witness}
v_n
:=
\frac{1}{n}
\sum_{t\in \mathcal I_1}
\hat\phi_t^{(1)}
\in \cH_{\cY}.
\end{equation}

To avoid the degenerate event \(v_n=0\), fix once and for all a unit vector \(u_0\in \cH_{\cY}\), and define the tie-broken pilot direction
\begin{equation}
\label{eq:pilot_direction}
w_n
:=
\begin{cases}
v_n/\|v_n\|_{\cH_{\cY}}, & v_n\neq 0,\\
u_0, & v_n=0.
\end{cases}
\end{equation}
By construction, \(\|w_n\|_{\cH_{\cY}}=1\) almost surely.

The second fold is then reduced to a scalar problem through the projected score
\begin{equation}
\label{eq:projected_score}
\xi_t(w_n)
:=
\left\langle
w_n,\hat\phi_t^{(2)}
\right\rangle_{\cH_{\cY}},
\qquad t\in \mathcal I_2.
\end{equation}

The next lemma identifies the scalar target induced by the projection.

\begin{lem}[Projected doubly robust identity]
\label{lem:projected_dr_identity}
For every \(t\in \mathcal I_2\),
\begin{equation}
\label{eq:projected_target_identity}
\E\!\left[
\xi_t(w_n)
\,\middle|\,
\cF_{t-1}
\right]
=
\left\langle
w_n,\Psi(a,a')
\right\rangle_{\cH_{\cY}}
=:
\theta_n(a,a').
\end{equation}
In particular, under \(H_0:\eta(a)=\eta(a')\),
\[
\E\!\left[
\xi_t(w_n)
\,\middle|\,
\cF_{t-1}
\right]
=0.
\]
\end{lem}

\subsection{Sequential standard deviation estimation}

For \(t\in \mathcal I_2\), define the oracle conditional standard deviation of the projected score by
\begin{equation}
\label{eq:oracle_projected_variance}
\sigma_{0,t}(w_n)
:=
\Var\!\left(
\xi_t(w_n)
\,\middle|\,
\cF_{t-1}
\right)^{1/2}.
\end{equation}

We estimate \(\sigma_{0,t}(w_n)\) using only observations from the past of the inferential fold. The key point is that a past observation \(X_s, A_s, Y_s\) is re-evaluated under the current policy \(\pi_t\) but with its own historical nuisance estimate \(\widehat\mu^{(2)}_{s-1}\), as noted in \citep{bibaut2021post}.

Fix \(t\in \mathcal I_2\), and let
\[
S_t := \{\, s\in \mathcal I_2 : s<t \,\},
\qquad
m_t := |S_t|.
\]
For every \(s\in S_t\), define the history-recycled RKHS score
\begin{equation}
\label{eq:phi_st_projected_section5}
\begin{aligned}
\check\phi_{s,t}^{(2)}
:= {} &
D'(\pi_t,\widehat\mu^{(2)}_{s-1},a)(X_s,A_s,Y_s) \\
&\quad
-
D'(\pi_t,\widehat\mu^{(2)}_{s-1},a')(X_s,A_s,Y_s)
\in \cH_{\cY}.
\end{aligned}
\end{equation}
Note that the \(s\)th observation is paired with the historical nuisance \(\widehat\mu^{(2)}_{s-1}\) and not with the current nuisance \(\widehat\mu^{(2)}_{t-1}\). We correct the discrepancy between the logging law at time \(s\) and the evaluation law at time \(t\) using the importance ratio
\begin{equation}
\label{eq:w_st_projected}
w_{s,t}
:=
\frac{\pi_t(A_s\mid X_s)}{\pi_s(A_s\mid X_s)}.
\end{equation}
The corresponding projected score is
\begin{equation}
\label{eq:xi_st_projected}
\check\xi_{s,t}(w_n)
:=
\left\langle
w_n,\check\phi_{s,t}^{(2)}
\right\rangle_{\cH_{\cY}}.
\end{equation}

Conditionally on \(\cF_{t-1}\), the pair \((\pi_t,w_n)\) is fixed, and the oracle projected moments are
\begin{align}
\label{eq:projected_oracle_moments}
M_{1,t}(w_n)
&:=
\E\!\left[
\xi_t(w_n)
\,\middle|\,
\cF_{t-1}
\right]
=
\theta_n(a,a'),
\\
M_{2,t}(w_n)
&:=
\E\!\left[
\xi_t(w_n)^2
\,\middle|\,
\cF_{t-1}
\right].
\end{align}
By definition,
\begin{equation}
\label{eq:projected_variance_decomposition}
\sigma_{0,t}^2(w_n)
=
M_{2,t}(w_n)-M_{1,t}(w_n)^2.
\end{equation}

For \(t=n+1\), there is no past data in the inferential fold. We therefore set
\begin{equation*}
\label{eq:first_inferential_variance_convention}
\widehat M_{1,n+1}(w_n):=0,
\
\widehat M_{2,n+1}(w_n):=1,
\
\widehat\sigma_{n+1}^2(w_n):=1.
\end{equation*}
This single convention is asymptotically negligible.

For \(t\ge n+2\), we estimate the moments by the importance-weighted empirical averages
\begin{align}
\label{eq:hatM1_projected}
\widehat M_{1,t}(w_n)
&:=
\frac{1}{m_t}
\sum_{s\in S_t}
w_{s,t}\,\check\xi_{s,t}(w_n),
\\
\label{eq:hatM2_projected}
\widehat M_{2,t}(w_n)
&:=
\frac{1}{m_t}
\sum_{s\in S_t}
w_{s,t}\,\check\xi_{s,t}(w_n)^2.
\end{align}
The plug-in projected conditional variance estimator is then
\begin{equation}
\label{eq:projected_variance_estimator}
\widehat\sigma_t^2(w_n)
:=
\Big(
\widehat M_{2,t}(w_n)-\widehat M_{1,t}(w_n)^2
\Big)_+,
\qquad t\ge n+2,
\end{equation}
where \((x)_+ := \max(x,0)\). All quantities above are \(\cF_{t-1}\)-measurable, so the resulting statistic remains fully predictable on the inferential fold.

To avoid division by zero in finite samples, let \((\epsilon_n)_{n\ge1}\) be a deterministic clip sequence with \(\epsilon_n\downarrow 0\), and define the clipped feasible standard deviation
\begin{equation}
\label{eq:clipped_sd}
\widetilde\sigma_t(w_n)
:=
\widehat\sigma_t(w_n)\vee \epsilon_n.
\end{equation}

Finally, the adaptive doubly robust kernel treatment effect test statistic is
\begin{equation}
\label{eq:final_test_stat}
T_n(w_n)
:=
\frac{1}{\sqrt n}
\sum_{t\in \mathcal I_2}
\widetilde\sigma_t(w_n)^{-1}\,\xi_t(w_n).
\tag{ADR-KTE}
\end{equation}

\subsection{Null asymptotic normality}
\label{section:projected_null_clt}

We now state the null asymptotic normality theorem. We now state mild assumptions on the adaptive data collection.

\begin{assum}[Exploration floor]
\label{assumption:exploration_floor}
There exist \(c_\pi>0\) and \(\alpha\in[0,1)\) such that
\begin{equation}
\label{eq:exploration_floor}
\pi_t(b\mid x)\ge c_\pi\,t^{-\alpha}
\end{equation}
for every \(t\ge1\), every \(b\in\cA\), and \(P_X\)-almost every \(x\in\cX\).
\end{assum}

The assumption above is milder than assuming lower bounded propensities, and verified by many bandit algorithms \citep{lattimore2020}. 

\begin{assum}[Asymptotically negligible clipping]
\label{assumption:negligible_clipping}
For the deterministic clip sequence \((\epsilon_n)\),
\begin{equation}
\label{eq:negligible_clipping}
\frac{1}{n}
\sum_{t\in\mathcal I_2}
\mathbbm 1\!\left\{
\sigma_{0,t}(w_n)\le 2\epsilon_n
\right\}
\xrightarrow[]{p}0.
\end{equation}
\end{assum}

\begin{thm}[Null asymptotic normality of the adaptive test]
\label{thm:projected_test_clt}
Assume Assumptions~\ref{assum:selection_observables},
\ref{assumption:bounded_kernel},
\eqref{eq:nuisance_envelope},
\ref{assumption:exploration_floor},
\ref{assumption:negligible_clipping}, and the assumptions of Theorem~\ref{thm:projected_variance_consistency}. Let \(\nu(\alpha,\beta,p)>0\) denote the rate constant from Theorem~\ref{thm:projected_variance_consistency}, and choose
\begin{equation*}
\label{eq:clip_rate_choice}
\epsilon_n = n^{-\rho}
\,\ \text{with} \,\
0<\rho<
\min\!\left\{
\frac{1-3\alpha}{4},
\frac{\nu(\alpha,\beta,p)}{2}
\right\}.
\end{equation*}
Then the statistic \(T_n(w_n)\) satisfies
\[
T_n(w_n)
\ \xRightarrow[]{d}\ \mathcal N(0,1)
\qquad
\text{under } H_0:\eta(a)=\eta(a').
\]
\end{thm}

The proof is deferred to Appendix~\ref{app:analysis_projected_test}. The next section is devoted to constructing \(\widehat\sigma_t(w_n)\) and proving the required average consistency of the feasible predictable variances by reducing the RKHS problem to a projected scalar variance-estimation problem.

\section{SEQUENTIAL CONDITIONAL STANDARD DEVIATION ESTIMATION}
\label{section:variance-estimation}

The statistic \(T_n(w_n)\) depends on the predictable conditional standard deviation of the projected score \(\xi_t(w_n)\). We now study the estimator introduced above. The key point is that after conditioning on the pilot fold, the projection direction \(w_n\) is fixed and the inferential-fold problem becomes scalar. 

\subsection{Sufficient conditions for projected variance consistency}

We first require a rate of convergence of the inferential-fold nuisance sequence to a fixed limit.

\begin{assum}[Nuisance stabilization rate]
\label{assumption:nuisance_rate}
There exist \(\mu_\infty:\cA\times\cX\to \cH_{\cY}\) and \(\beta>0\) such that
\begin{equation}
\label{eq:nuisance_rate}
\big\|
\widehat\mu^{(2)}_{t-1}-\mu_\infty
\big\|_{L_2(P_X\times \mu_{\cA};\,\cH_{\cY})}
=
O(t^{-\beta})
\qquad \text{a.s.}
\end{equation}
\end{assum}

Note that above, we only require the nuisance estimate to converge to a fix limit which may be mispecified. This is allowed by our use of DR RKHS scores. Next, we assume the condition below which will quantify the complexity of the logging policies through their density ratios with respect a reference policy \(\pi_{\mathrm{ref}}\). 

\begin{assum}[Logging-policy class regularity]
\label{assumption:policy_complexity}
There exist a reference policy $\pi_{\mathrm{ref}}$ and constants \(G<\infty\) and \(p>0\) such that
\begin{equation}
\label{eq:policy_ratio_bound}
\sup_{g\in\mathcal G_{\mathrm{ref}}}\|g\|_\infty \le G,
\end{equation}
and
\begin{equation}
\label{eq:policy_entropy_bound}
\log N_{[]}\!\left(\varepsilon,\mathcal G_{\mathrm{ref}},\|\cdot\|_{2,\mathrm{ref}}\right)
\lesssim \varepsilon^{-p}.
\end{equation}

where we define $\mathcal G_{\mathrm{ref}}
:=
\left\{
\frac{\pi}{\pi_{\mathrm{ref}}}
:\,
\pi\in\Pi
\right\}$,
and equip this class with the norm
\[
\|h\|_{2,\mathrm{ref}}^2
:=
\E_{X\sim P_X}
\!\left[
\int h(b,X)^2\,\pi_{\mathrm{ref}}(b\mid X)\,\diff\mu_{\cA}(b)
\right].
\]
\end{assum}

Finally, for consistency of the variance estimator we require the same strengthened exploration floor as in \citet{bibaut2021post}.

\begin{assum}[Strengthened exploration floor]
\label{assumption:exploration_floor_strong}
Assumption~\ref{assumption:exploration_floor} holds with exponent
\begin{equation}
\label{eq:alpha_star}
\alpha
<
\alpha_\star(\beta,p)
:=
\min\!\left\{
\frac{1}{3+p},
\frac{1}{1+2p},
\beta
\right\}.
\end{equation}
\end{assum}

\begin{thm}[Average consistency of the projected conditional standard deviation estimator]
\label{thm:projected_variance_consistency}
Assume \textup{\ref{assum:selection_observables}},
\textup{\ref{assumption:bounded_kernel}},
\eqref{eq:nuisance_envelope},
\textup{\ref{assumption:nuisance_rate}},
\textup{\ref{assumption:policy_complexity}}, and
\textup{\ref{assumption:exploration_floor_strong}}.
Then there exists a constant \(\nu(\alpha,\beta,p)>0\) such that
\begin{equation}
\label{eq:average_M1_consistency}
\frac{1}{n}
\sum_{t\in\mathcal I_2}
\big|
\widehat M_{1,t}(w_n)-M_{1,t}(w_n)
\big|
=
O_{\mathrm{a.s.}}\!\big(n^{-\nu(\alpha,\beta,p)}\big),
\end{equation}
\begin{equation}
\label{eq:average_M2_consistency}
\frac{1}{n}
\sum_{t\in\mathcal I_2}
\big|
\widehat M_{2,t}(w_n)-M_{2,t}(w_n)
\big|
=
O_{\mathrm{a.s.}}\!\big(n^{-\nu(\alpha,\beta,p)}\big),
\end{equation}
and consequently
\begin{equation}
\label{eq:average_sigma_consistency}
\frac{1}{n}
\sum_{t\in\mathcal I_2}
\big|
\widehat\sigma_t^2(w_n)-\sigma_{0,t}^2(w_n)
\big|
=
O_{\mathrm{a.s.}}\!\big(n^{-\nu(\alpha,\beta,p)}\big).
\end{equation}
\end{thm}

\begin{cor}[Average inverse-standard-deviation consistency]
\label{assumption:average_inverse_sd_consistency}
Assume the conditions of Theorem~\ref{thm:projected_variance_consistency} and Assumption~\ref{assumption:negligible_clipping}. If
\[
\epsilon_n=n^{-\rho}
\qquad\text{with}\qquad
0<\rho<\frac{\nu(\alpha,\beta,p)}{2},
\]
then
\begin{equation}
\label{eq:average_inverse_sd_consistency}
\frac{1}{n}
\sum_{t\in\mathcal I_2}
\left|
\frac{\sigma_{0,t}^2(w_n)}{\widetilde\sigma_t^2(w_n)}-1
\right|
\xrightarrow[]{p}0.
\end{equation}
In particular,
\begin{equation}
\label{eq:predictable_variation_consistency}
\frac{1}{n}
\sum_{t\in\mathcal I_2}
\frac{\sigma_{0,t}^2(w_n)}{\widetilde\sigma_t^2(w_n)}
\xrightarrow[]{p}1.
\end{equation}
\end{cor}

The proof is deferred to Appendix~\ref{app:projected_variance_consistency}. The proof is purely scalar after conditioning on the pilot fold and follows the a similar decomposition as that of the variance-estimation proof of \citet{bibaut2021post}, now applied to the projected outcome.

\section{PRACTICAL PROCEDURE AND POWER}
\label{section:practical_test_power}

Sections~\ref{section:projected_null_clt} and \ref{section:variance-estimation}
separate the analysis into two parts: a null asymptotic normality result for the normalized statistic \(T_n(w_n)\), and a projected variance-estimation theorem proving the predictable-variation consistency required by that CLT. We now combine these pieces and study the behavior of the test under fixed alternatives.

\subsection{Practical adaptive KTE test}

The practical procedure of the \eqref{eq:final_test_stat} test is summarized in
Algorithm~\ref{alg:projected_kte_test} (more details in Appendix \ref{appendix:ClosedFormDerivation}). It uses a chronological pilot fold to
construct an RKHS witness and an inferential fold to compute the normalized
projected score.

\begin{cor}[Asymptotic validity of the adaptive KTE test]
\label{cor:projected_test_validity}
Under the assumptions of Theorem~\ref{thm:projected_test_clt},
\[
T_n(w_n)\ \xRightarrow[]{d}\ \mathcal N(0,1)
\qquad
\text{under } H_0:\eta(a)=\eta(a').
\]
Consequently, the rejection rule
\[
\phi_n
=
\mathbbm{1}\!\left\{
T_n(w_n)>z_{1-\gamma}
\right\}
\]
has asymptotic level \(\gamma\).
\end{cor}

\begin{algorithm}[t]
\footnotesize
\setlength{\abovedisplayskip}{2pt}
\setlength{\belowdisplayskip}{2pt}
\setlength{\abovecaptionskip}{2pt}
\setlength{\textfloatsep}{4pt}
\caption{Adaptive KTE test (ADR-KTE)}
\label{alg:projected_kte_test}
\begin{algorithmic}[1]
\State \textbf{Input:} adaptive data \(\mathcal D_T\), policies \(\{\pi_t\}_{t=1}^T\), target actions \((a,a')\), significance level \(\gamma\)
\State Split \(\{1,\dots,T\}\) chronologically into \(\mathcal I_1=\{1,\dots,n\}\) and \(\mathcal I_2=\{n+1,\dots,2n\}\)
\For{each \(t\in \mathcal I_1\)}
    \State construct a predictable nuisance estimate \(\widehat\mu^{(1)}_{t-1}\)
    \State compute the pilot RKHS score \(\hat\phi_t^{(1)}\) from \eqref{eq:phi_t_fold}
\EndFor
\State Form the pilot witness \(v_n=\frac{1}{n}\sum_{t\in\mathcal I_1}\hat\phi_t^{(1)}\)
\State Set the tie-broken pilot direction
\[
w_n=\begin{cases}
v_n/\|v_n\|_{\cH_{\cY}}, & v_n\neq 0,\\
u_0, & v_n=0,
\end{cases}
\]
where \(u_0\in\cH_{\cY}\) is fixed with \(\|u_0\|_{\cH_{\cY}}=1\)
\For{each \(t\in \mathcal I_2\)}
    \State construct a predictable nuisance estimate \(\widehat\mu^{(2)}_{t-1}\)
    \State compute \(\xi_t(w_n)=\langle w_n,\hat\phi_t^{(2)}\rangle_{\cH_{\cY}}\)
    \If{\(t=n+1\)}
        \State set \(\widehat\sigma_t(w_n)=1\)
    \Else
        \State let \(S_t=\{s\in\mathcal I_2:s<t\}\)
        \For{each \(s\in S_t\)}
            \State compute \(w_{s,t}=\pi_t(A_s\mid X_s)/\pi_s(A_s\mid X_s)\)
            \State compute \(\check\xi_{s,t}(w_n)=\langle w_n,\check\phi_{s,t}^{(2)}\rangle_{\cH_{\cY}}\)
        \EndFor
        \State compute \(\widehat M_{1,t}(w_n)\) and \(\widehat M_{2,t}(w_n)\) from \eqref{eq:hatM1_projected}--\eqref{eq:hatM2_projected}
        \State set \(\widehat\sigma_t^2(w_n)=\big(\widehat M_{2,t}(w_n)-\widehat M_{1,t}(w_n)^2\big)_+\)
        \State set \(\widehat\sigma_t(w_n)=\sqrt{\widehat\sigma_t^2(w_n)}\)
    \EndIf
    \State set \(\widetilde\sigma_t(w_n)=\widehat\sigma_t(w_n)\vee \epsilon_n\)
\EndFor
\State Set \(T_n(w_n)=\frac{1}{\sqrt n}\sum_{t\in\mathcal I_2}\widetilde\sigma_t(w_n)^{-1}\xi_t(w_n)\)
\State \textbf{Reject \(H_0\)} if \(T_n(w_n)>z_{1-\gamma}\)
\end{algorithmic}
\end{algorithm}

Corollary~\ref{cor:projected_test_validity} gives the final null calibration guarantee. The statistic is one-dimensional, but its signal is inherited from the RKHS-valued effect \(\Psi(a,a')\) through the pilot witness \(v_n\).

\subsection{Witness consistency}

We next show that the pilot witness consistently estimates \(\Psi(a,a')\). This
result is separate from the variance-estimation analysis and requires only the
boundedness of the kernel, the nuisance envelope
\eqref{eq:nuisance_envelope}, and the exploration floor.

\begin{thm}[Pilot witness consistency]
\label{thm:pilot_witness_consistency}
Under Assumptions~\ref{assum:selection_observables},
\ref{assumption:bounded_kernel},
\ref{assumption:exploration_floor}, and the nuisance envelope
\eqref{eq:nuisance_envelope}, the pilot witness
\eqref{eq:pilot_witness} satisfies
\begin{equation}
\label{eq:pilot_l2_consistency}
\E\!\left[
\|v_n-\Psi(a,a')\|_{\cH_{\cY}}^2
\right]
\lesssim n^{\alpha-1}.
\end{equation}
In particular, if \(\alpha<1\), then
\[
v_n \xrightarrow[]{L_2(\cH_{\cY})} \Psi(a,a'),
\qquad
v_n \xrightarrow[]{p} \Psi(a,a').
\]
\end{thm}

The proof is deferred to Appendix~\ref{app:witness_power}.

\begin{cor}[Consistency of the pilot direction]
\label{cor:pilot_direction_consistency}
Assume the conditions of Theorem~\ref{thm:pilot_witness_consistency} and let \(\Psi(a,a')\neq 0\). Then
\[
w_n
\xrightarrow[]{p}
\frac{\Psi(a,a')}{\|\Psi(a,a')\|_{\cH_{\cY}}}.
\]
\end{cor}

The proof is deferred to Appendix~\ref{app:witness_power}.

Theorem~\ref{thm:pilot_witness_consistency} is the basic reason the projected
statistic remains meaningful under fixed alternatives. The witness is learned
from a raw RKHS average. Stabilization is only needed on the inferential fold,
where the goal is valid asymptotic normality of the final normalized statistic.

\subsection{Consistency under fixed alternatives}

We now turn to the behavior of the test under \(H_1\).

\begin{cor}[Consistency under fixed alternatives]
\label{cor:projected_test_power}
Assume the conditions of Corollary~\ref{cor:projected_test_validity} and let \(\Psi(a,a')\neq 0\). Then
\[
T_n(w_n)\xrightarrow[]{p} +\infty,
\]
and therefore the rejection rule in
Corollary~\ref{cor:projected_test_validity} is consistent.
\end{cor}

\begin{rem}[Deterministic limit direction under \(H_1\)]
\label{rem:deterministic_limit_direction}
When \(\Psi(a,a')\neq 0\), write
\[
w_\star:=\frac{\Psi(a,a')}{\|\Psi(a,a')\|_{\cH_{\cY}}}.
\]
The power proof uses only that \(w_n\xrightarrow[]{p} w_\star\). No uniform lower variance bound over random pilot directions is imposed.
\end{rem}

The proof is deferred to Appendix~\ref{app:witness_power}. Corollary~\ref{cor:projected_test_power} shows that the test remains sensitive
to fixed distributional alternatives. The chronology of the method is therefore
simple: the first fold learns the witness, the second fold estimates the conditional standard deviation of the projected score using the projected CADR construction, and the resulting normalized statistic is asymptotically valid under the null and consistent under fixed alternatives.

\begin{figure*}[ht]
  \centering
  \includegraphics[width=0.89\textwidth]{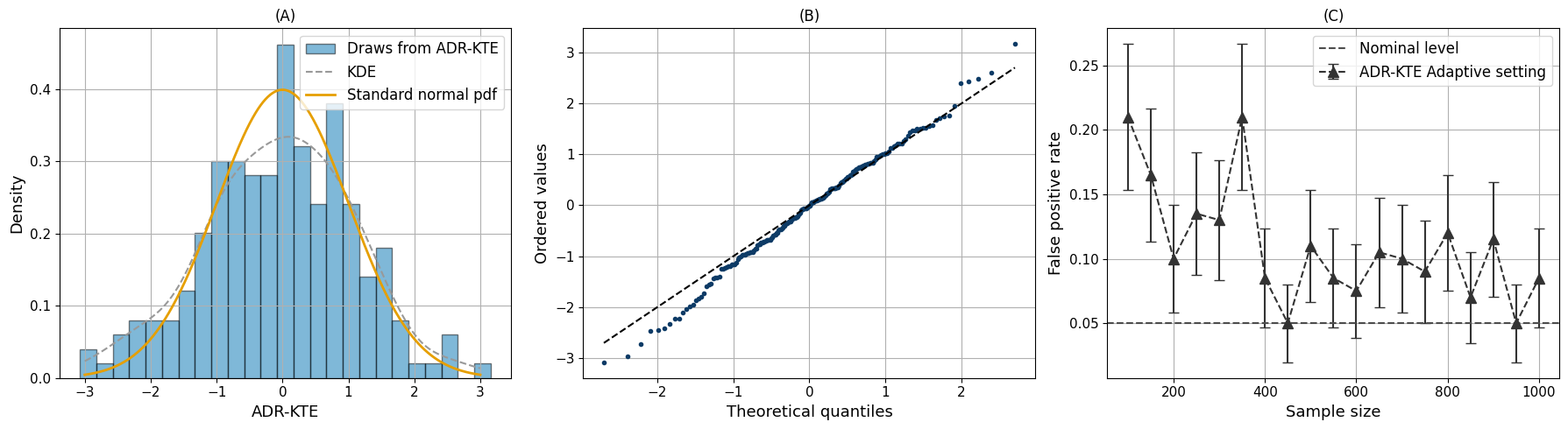}
\caption{Illustration of $200$ simulations of \textsc{ADR-KTE} under the null in the adaptive setting with $T=950$: (A) Histogram with KDE and standard normal pdf, (B) Normal Q-Q plot, (C) False positives against sample sizes. The results show approximate Gaussian behaviour and controlled type-I error. }
\label{fig:null_adaptive}
\end{figure*}

\section{NUMERICAL SIMULATIONS}
\label{section:experiments}

In this section, we study the empirical calibration and power of our proposed test
\textsc{ADR-KTE} under \emph{adaptive} data collection. We observe a stream
\(\{(X_t,A_t,Y_t)\}_{t=1}^T\) generated by a bandit-style logging
policy \(\pi_t(\cdot\mid X_t)\). We evaluate both
calibration (Scenario~I) and power (Scenarios~II--IV) at a significance level of
\(\alpha=0.05\). Additional details and results appear in
Appendix~\ref{app:additional_experiments}. Our implementation is available on GitHub.\footnote{\url{https://github.com/houssamzenati/adaptive-KTE}} \looseness=-1

\textbf{Adaptive data collection.}
Actions follow an $\varepsilon$-greedy contextual bandit with per–arm online ridge. 
At time $t$, with features $Z_t=[1,X_t]$, each arm $a$ has 
$S_a=\lambda I+\sum_{s\le t:A_s=a}Z_sZ_s^\top$, 
$b_a=\sum_{s\le t:A_s=a}Z_sY_s$, 
$\hat\theta_a=S_a^\dagger b_a$, 
and score $q_a(t)=Z_t^\top\hat\theta_a$. 
The propensity is\looseness=-1
\[ \pi_t(1\mid X_t)= \begin{cases} 1-\tfrac{1}{2}\varepsilon_t, & q_1(t)>q_0(t),\\ \tfrac{1}{2}\varepsilon_t, & q_1(t)<q_0(t),\\ \tfrac{1}{2}, & \text{otherwise,} \end{cases} \]
with $\varepsilon_0\!\in\!(0,1)$, $\varepsilon_{\min}\!>\!0$, $\beta\!\in\!(0,1]$. We sample $A_t\!\sim\!\pi_t$, observe $Y_t$, and store $\pi_t(A_t\mid X_t)$. For sample-splitting we use non-overlapping time folds: by default an \emph{alternating} 
split ($\mathcal{I}_0=\{t\ \text{odd}\}$, $\mathcal{I}_1=\{t\ \text{even}\}$). Each fold is evaluated in temporal order so all nuisance weights remain predictable.

\textbf{Baselines.}
We compare to two adaptive inference methods: \textbf{CADR} \citep{bibaut2021post}, stabilizes the DR score with history-measurable weights that estimate its conditional variance from past data, yielding a martingale CLT, and \textbf{AW-AIPW} \citep{hadad2021confidence}, enforces deterministic quadratic variation in adaptive experiments by reweighting AIPW scores with variance-stabilizing allocations, guaranteeing asymptotic normality. \emph{Both are scalar, targeting mean effects} (i.e., contrasts of $\mathbb{E}[Y(a)]$), whereas our ADR-KTE directly targets \emph{the full outcome distribution} via RKHS mean embeddings. We use the authors’ open-source implementations and fit the regression nuisances with kernel ridge regressions; details are in Appendix~\ref{app:additional_experiments}.

\subsection{Synthetic data.}
\label{sec:SyntheticDataNumExperiment}
We adapt the synthetic designs of \citet{martinez2023} to the adaptive setting by replacing i.i.d.\ assignment with an $\epsilon$-greedy policy $\{\pi_t\}$ as described above. Each replicate simulates covariates $X \in \mathbb{R}^5$, draws $T$ rounds under $\pi_t$. The potential outcomes are defined as $Y_t(A_t) = \cos(\beta^\top X_t) + \Delta(s) \mathbf{1}(A_t = 1) +  \epsilon_{t},$ with $\beta = (0.1, 0.2, 0.3, 0.4, 0.5)^\top$, independent noises $\epsilon_{t} \sim \mathcal{N}(0,0.5)$, and the shift random variable $\Delta(s)$ varied to match each scenario $s$. Four scenarios are considered for $\Delta(s)$: (I) no effect; (II) mean shift only; (III--IV) higher-moment changes at equal means. Additional details and other forms of potential outcome function $Y_t(A_t)$ experimented are given in Appendix~\ref{app:additional_experiments}.

\begin{figure*}[ht]
  \centering
\includegraphics[width=0.89\linewidth]{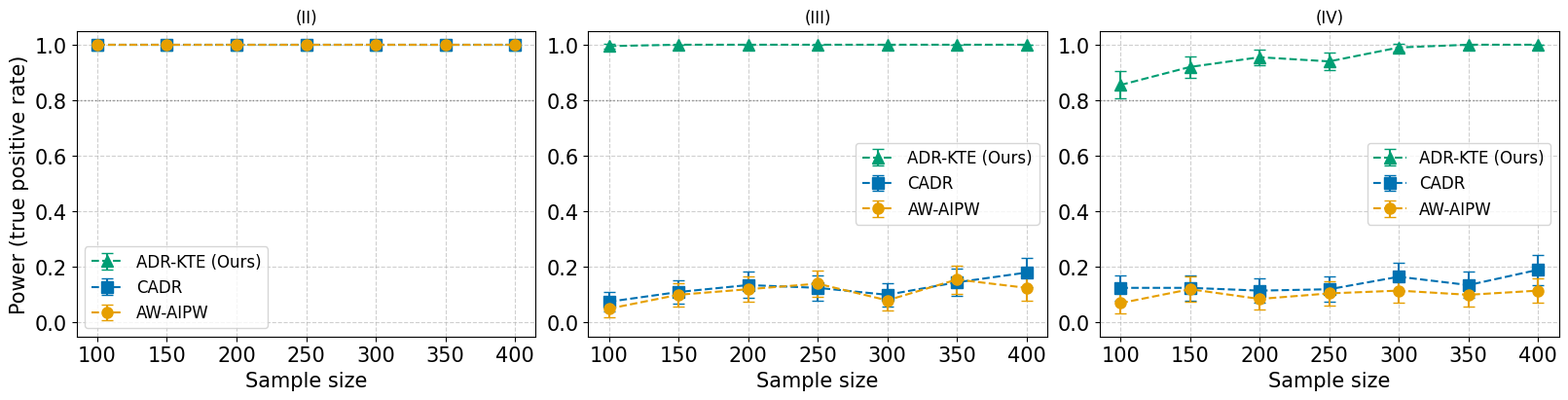}
\caption{True positive rates ($200$ simulations, Scenarios II–IV). Mean-focused baselines (CADR/AW-AIPW) achieve matching performance on II; ADR-KTE shows markedly higher power on III–IV (higher-moment shifts).}
\label{fig:adaptive_power}
\end{figure*}

In Scenario~I, \textsc{ADR-KTE} is well calibrated (see the empirical histogram, QQ-plot and false positive rate in Figure~\ref{fig:null_adaptive}). Across Scenarios~II–IV (Figure~\ref{fig:adaptive_power}), it attains high power for both mean and higher-moment shifts. By contrast, ATE-focused baselines (CADR, AW-AIPW) match only under mean shifts (II) and fail under purely distributional changes (III–IV).



\subsection{IHDP dataset}
We evaluate our method on the Infant Health and Development Program (IHDP) data \citep{Hill01012011}, following the same design as in \citep{martinez2023}: after removing missing rows we retain 908 units with 18 covariates (9 continuous, 9 categorical). In our experiment, treatments are assigned adaptively via the $\epsilon$-greedy policy described earlier. The outcome construction mirrors the simulation design of previous Scenarios (I)–(IV), where potential outcomes are similarly defined as $Y_t(A_t) = \cos(\beta^\top X_t) + \Delta(s) \mathbf{1}(A_t = 1) +  \epsilon_{t},$ with $\beta = (1,\dots,1)^\top$, independent Gaussian noises $\epsilon_{t} \sim \mathcal{N}(0,0.5)$, and the shift random variable $\Delta(s)$ varied to match each scenario $s$ (zero under the null, mean shift in II, equal-mean distributional changes in III–IV). Full implementation details are provided in Appendix~\ref{app:additional_experiments}.\looseness=-1



Table \ref{table:IHDP_results_main} reports true positive rates (mean ± standard error). \textsc{ADR-KTE} achieves near-perfect power across Scenarios II–IV, illustrating the benefits of our distributional kernel test under adaptivity. Conversely, CADR and AW-AIPW succeed only on the mean shift (II), largely failing (rejection rates $\approx\alpha$) under equal-mean distributional shifts (III–IV).

\begin{table}[h]
\centering
\caption{True positive rates (mean $\pm$ se) for IHDP on $200$ simulations and a sample size $T = 908$. 
}
\label{table:IHDP_results_main}
\begin{adjustbox}{width=\columnwidth}
\begin{tabular}{lrrr}
\toprule
 & II & III & IV \\
\midrule
ADR-KTE & $\mathbf{1.0 \pm 0.0}$ & $\mathbf{1.0 \pm 0.0}$   & $\mathbf{1.0 \pm 0.0}$   \\
CADR      & $1.0 \pm 0.0$ & $0.075\pm 0.04$ & $0.085 \pm 0.04$ \\
AW-AIPW   & $1.0 \pm 0.0$ & $0.07 \pm 0.04$ & $0.045 \pm 0.03$ \\
\bottomrule
\end{tabular}
\end{adjustbox}
\end{table}

\subsection{dSprite dataset}

We evaluate our kernel test on the \emph{dSprites} dataset \citep{dsprites17} with structured outcomes $Y\in\mathbb{R}^{64\times64}$. Contexts $X\sim \mathrm{Unif}([0,1]^2)$ are mapped to images $Y$ by a deterministic renderer $g(X,A)$ that places a white heart shape in a black image based on $X,A$. We study two regimes: Scenario~I (null), where both treatments induce the same image distribution, and Scenario~IV (shift), where $A=1$ translates the heart shape relative to $A=0$ (a spatial change with unchanged mean intensity). Logged data are collected by an \emph{adaptive} $\varepsilon$-greedy policy with per-arm online ridge. Our test, \textsc{ADR-KTE}, operates directly on flattened images. By contrast, baselines (\textsc{CADR} and \textsc{AW-AIPW}) require scalar outcomes, forcing us to use the mean pixel per image, which inherently cannot detect the spatial shift in Scenario IV.\looseness=-1

\begin{table}[h]
\centering
\small
\caption{True positive rates (mean $\pm$ se) for dSprite on 200 simulations and a sample size of $T=1000$.}
\begin{tabular}{lll}
\toprule
 & I & IV \\
\midrule
ADR-KTE & \boldmath$0.04\,\pm\,0.03$\unboldmath & \boldmath$1.00\,\pm\,0.00$\unboldmath \\
CADR & $0.19\,\pm\,0.05$ & $0.19\,\pm\,0.05$ \\
AW-AIPW & $0.10\,\pm\,0.04$ & $0.10\,\pm\,0.04$ \\
\bottomrule
\end{tabular}
\label{tab:dsprites}
\end{table}

\textsc{ADR-KTE} shows near-nominal type-I error in Scenario~I and perfect power in Scenario~IV, detecting the spatial shift in the full image distribution. In contrast, \textsc{CADR} and \textsc{AW-AIPW} (fed only the mean pixel) exhibit non-trivial false positives under the null and no power in the shift scenario, underscoring the value of testing for structured outcomes.

\subsection{Comparison with non-adaptive baselines}
\label{sec:iid_vs_adaptive_main}

To complement the comparison with adaptive baselines, we also evaluate against kernel procedures designed for i.i.d.\ data, namely \textsc{DR-xKTE} \citep{martinez2023} and \textsc{KTE} \citep{muandet2021counterfactual}. Appendix~\ref{app:iid_comparison} reports results under both i.i.d.\ and adaptive designs. Under i.i.d.\ sampling, all three methods are correctly calibrated and attain full power. Under an $\varepsilon$-greedy adaptive design, which maintains persistent exploration and falls in a stable adaptive regime \citep{laiweiaos1982}, the picture changes: \textsc{DR-xKTE} becomes anti-conservative under the null, \textsc{KTE} remains roughly calibrated but suffers a substantial loss of power, and \textsc{ADR-KTE} achieves the strongest calibration-power tradeoff, with null rejection rates moving toward the nominal level as $T$ increases and power close to one. These results show that even under a stable adaptive design, non-adaptive kernel tests are not reliable without explicit variance control.

\section{DISCUSSION}
\label{section:discussion}

We introduced \textsc{ADR-KTE}, a kernel-based test for distributional treatment effects under adaptive data collection. Our method uses a learn-then-test construction: a first chronological fold estimates a witness direction in the outcome RKHS, and inference is then performed on the associated projected score. The resulting scalar statistic remains sensitive to general distributional differences, including shifts beyond the mean. Recent work \citep{shen2026} shows that, under a stability condition \citep{laiweiaos1982}, estimators that are asymptotically normal in the i.i.d.\ setting may remain so under adaptive designs without additional stabilization. A key open question is whether a similar principle holds for kernel-based distributional estimators beyond the projected setting considered here. Other promising directions include conditional effects and richer embedding regressors.

\subsubsection*{Acknowledgements}
Houssam Zenati, Bariscan Bozkurt, and Arthur Gretton are supported by the Gatsby Charitable Foundation. The authors thank Zikai Shen for relevant remarks on an earlier version of this work. The authors also thank the AISTATS 2026 reviewers and the Area Chair for their constructive feedback and helpful discussions during the review process.

\bibliography{references}

@article{SriGreFukLanetal10,
  Author =	 {B. Sriperumbudur and A. Gretton and K. Fukumizu and
                  B. Sch{\"o}lkopf and G. Lanckriet},
  Journal =	 {Journal of Machine Learning Research},
  Pages =	 {1517-1561},
  Title =	 {Hilbert Space Embeddings and Metrics on Probability
                  Measures},
  Volume =	 11,
  Year =	 2010
}

@book{berlinet2011reproducing,
  title={Reproducing kernel Hilbert spaces in probability and statistics},
  author={Berlinet, Alain and Thomas-Agnan, Christine},
  year={2011},
  publisher={Springer Science \& Business Media}
}

@book{hall1980,
  title={Martingale limit theory and its application},
  author={Hall, Peter and Heyde, Christopher C},
  year={1980},
  publisher={Academic press}
}

@article{bibaut2021post,
  title={Post-contextual-bandit inference},
  author={Bibaut, Aur{\'e}lien and Dimakopoulou, Maria and Kallus, Nathan and Chambaz, Antoine and van Der Laan, Mark},
  journal={Advances in neural information processing systems},
  volume={34},
  pages={28548--28559},
  year={2021}
}

@inproceedings{martinez2023,
 author = {Martinez Taboada, Diego and Ramdas, Aaditya and Kennedy, Edward},
 booktitle = {Advances in Neural Information Processing Systems},
 pages = {59924--59952},
 title = {An Efficient Doubly-Robust Test for the Kernel Treatment Effect},
 volume = {36},
 year = {2023}
}

@article{
fawkes2024doubly,
title={Doubly Robust Kernel Statistics for Testing Distributional Treatment Effects},
author={Jake Fawkes and Robert Hu and Robin J. Evans and Dino Sejdinovic},
journal={Transactions on Machine Learning Research},
year={2024},
}

@article{luedtke2024,
author = {Alex Luedtke and Incheoul Chung},
title = {{One-step estimation of differentiable Hilbert-valued parameters}},
volume = {52},
journal = {The Annals of Statistics},
number = {4},
pages = {1534 -- 1563},
year = {2024},
}

@inproceedings{huang2021,
 author = {Huang, Audrey and Leqi, Liu and Lipton, Zachary and Azizzadenesheli, Kamyar},
 booktitle = {Advances in Neural Information Processing Systems},
 pages = {23714--23726},
 title = {Off-Policy Risk Assessment in Contextual Bandits},
 volume = {34},
 year = {2021}
}

@article{rockafellar2000optimization,
  title={Optimization of conditional value-at-risk},
  author={Rockafellar, R Tyrrell and Uryasev, Stanislav and others},
  journal={Journal of risk},
  volume={2},
  pages={21--42},
  year={2000},
}

@article{vaart1997weak,
  title={Weak convergence and empirical processes with applications to statistics},
  author={Vaart, AW van der and Wellner, Jon A},
  journal={Journal of the Royal Statistical Society-Series A Statistics in Society},
  volume={160},
  number={3},
  pages={596--608},
  year={1997},
}

@misc{dsprites17,
author = {Loic Matthey and Irina Higgins and Demis Hassabis and Alexander Lerchner},
title = {dSprites: Disentanglement testing Sprites dataset},
howpublished= {https://github.com/deepmind/dsprites-dataset/},
year = "2017",
}

@article{kim2024dimension,
  title={Dimension-agnostic inference using cross U-statistics},
  author={Kim, Ilmun and Ramdas, Aaditya},
  journal={Bernoulli},
  volume={30},
  number={1},
  pages={683--711},
  year={2024},
  publisher={Bernoulli Society for Mathematical Statistics and Probability}
}

@article{li2022optimal,
  title={Optimal rates for regularized conditional mean embedding learning},
  author={Li, Zhu and Meunier, Dimitri and Mollenhauer, Mattes and Gretton, Arthur},
  journal={Advances in Neural Information Processing Systems},
  volume={35},
  pages={4433--4445},
  year={2022}
}

@article{singh2024kernel,
  title={Kernel methods for causal functions: dose, heterogeneous and incremental response curves},
  author={Singh, Rahul and Xu, Liyuan and Gretton, Arthur},
  journal={Biometrika},
  volume={111},
  number={2},
  pages={497--516},
  year={2024},
  publisher={Oxford University Press}
}

@article{xu2023causal,
  title={Causal benchmark based on disentangled image dataset},
  author={Xu, Liyuan and Gretton, Arthur},
  year={2023}
}

@inproceedings{smola2007hilbert,
  title={A Hilbert space embedding for distributions},
  author={Smola, Alex and Gretton, Arthur and Song, Le and Sch{\"o}lkopf, Bernhard},
  booktitle={International conference on algorithmic learning theory},
  pages={13--31},
  year={2007},
  organization={Springer}
}

@article{rothe2010nonparametric,
  title={Nonparametric estimation of distributional policy effects},
  author={Rothe, Christoph},
  journal={Journal of Econometrics},
  volume={155},
  number={1},
  pages={56--70},
  year={2010},
}

@article{chernozhukov2013inference,
  title={Inference on counterfactual distributions},
  author={Chernozhukov, Victor and Fern{\'a}ndez-Val, Iv{\'a}n and Melly, Blaise},
  journal={Econometrica},
  volume={81},
  number={6},
  pages={2205--2268},
  year={2013},
}

@book{lattimore2020, place={Cambridge}, title={Bandit Algorithms}, publisher={Cambridge University Press}, author={Lattimore, Tor and Szepesvári, Csaba}, year={2020}}

@article{muandet2021counterfactual,
  title={Counterfactual mean embeddings},
  author={Muandet, Krikamol and Kanagawa, Motonobu and Saengkyongam, Sorawit and Marukatat, Sanparith},
  journal={Journal of Machine Learning Research},
  volume={22},
  number={162},
  pages={1--71},
  year={2021}
}

@article{gretton2012kernel,
  title={A kernel two-sample test},
  author={Gretton, Arthur and Borgwardt, Karsten M and Rasch, Malte J and Sch{\"o}lkopf, Bernhard and Smola, Alexander},
  journal={The Journal of Machine Learning Research},
  volume={13},
  number={1},
  pages={723--773},
  year={2012},
}

@inproceedings{dabney2018implicit,
  title={Implicit quantile networks for distributional reinforcement learning},
  author={Dabney, Will and Ostrovski, Georg and Silver, David and Munos, R{\'e}mi},
  booktitle={International conference on machine learning},
  pages={1096--1105},
  year={2018},
  organization={PMLR}
}

@inproceedings{park2021conditional,
  title={Conditional distributional treatment effect with kernel conditional mean embeddings and u-statistic regression},
  author={Park, Junhyung and Shalit, Uri and Sch{\"o}lkopf, Bernhard and Muandet, Krikamol},
  booktitle={International conference on machine learning},
  pages={8401--8412},
  year={2021},
}

@article{gartner2003survey,
  title={A survey of kernels for structured data},
  author={G{\"a}rtner, Thomas},
  journal={ACM SIGKDD explorations newsletter},
  volume={5},
  number={1},
  pages={49--58},
  year={2003},
}

@article{gretton2013introduction,
  title={Introduction to rkhs, and some simple kernel algorithms},
  author={Gretton, Arthur},
  journal={Adv. Top. Mach. Learn. Lecture Conducted from University College London},
  volume={16},
  number={5-3},
  pages={2},
  year={2013}
}

@InProceedings{kanagawa14,
  title = 	 {{Recovering Distributions from Gaussian RKHS Embeddings}},
  author = 	 {Kanagawa, Motonobu and Fukumizu, Kenji},
  booktitle = 	 {Proceedings of the Seventeenth International Conference on Artificial Intelligence and Statistics},
  year = 	 {2014},
  volume = 	 {33},
}

@inproceedings{song2009hilbert,
  title={Hilbert space embeddings of conditional distributions with applications to dynamical systems},
  author={Song, Le and Huang, Jonathan and Smola, Alex and Fukumizu, Kenji},
  booktitle={Proceedings of the 26th Annual International Conference on Machine Learning},
  pages={961--968},
  year={2009}
}

@article{shekhar2023,
  author  = {Shubhanshu Shekhar and Ilmun Kim and Aaditya Ramdas},
  title   = {A Permutation-Free Kernel Independence Test},
  journal = {Journal of Machine Learning Research},
  year    = {2023},
  volume  = {24},
  number  = {369},
  pages   = {1--68},
}

@book{vanderVaart1998asymptotic,
  author    = {van der Vaart, Aad W.},
  title     = {Asymptotic Statistics},
  series    = {Cambridge Series in Statistical and Probabilistic Mathematics},
  publisher = {Cambridge University Press},
  year      = {1998},
  doi       = {10.1017/CBO9780511802256}
}

@article{gretton12a,
  author  = {Arthur Gretton and Karsten M. Borgwardt and Malte J. Rasch and Bernhard Sch{{\"o}}lkopf and Alexander Smola},
  title   = {A Kernel Two-Sample Test},
  journal = {Journal of Machine Learning Research},
  year    = {2012},
  volume  = {13},
  number  = {25},
  pages   = {723-773},
}

@inproceedings{zenati2025cpme,
 author={Houssam Zenati and Bariscan Bozkurt and Arthur Gretton},
 title={Doubly-Robust Estimation of Counterfactual Policy Mean Embeddings}, 
  booktitle={Advances in Neural Information Processing Systems (NeurIPS)},
  year={2025}
}

@inproceedings{zenati22a,
  title={Efficient kernelized ucb for contextual bandits},
  author={Zenati, Houssam and Bietti, Alberto and Diemert, Eustache and Mairal, Julien and Martin, Matthieu and Gaillard, Pierre},
  booktitle={International Conference on Artificial Intelligence and Statistics (AISTATS)},
  year={2022},
}

@article{laiweiaos1982,
 author = {Tze Leung Lai and Ching Zong Wei},
 journal = {The Annals of Statistics},
 number = {1},
 pages = {154--166},
 publisher = {Institute of Mathematical Statistics},
 title = {Least Squares Estimates in Stochastic Regression Models with Applications to Identification and Control of Dynamic Systems},
 urldate = {2025-12-21},
 volume = {10},
 year = {1982}
}

@misc{shen2026,
      title={Efficient Inference after Directionally Stable Adaptive Experiments}, 
      author={Zikai Shen and Houssam Zenati and Nathan Kallus and Arthur Gretton and Koulik Khamaru and Aurélien Bibaut},
      year={2026},
      url={https://arxiv.org/abs/2602.21478}, 
}

@article{hadad2021confidence,
  title={Confidence intervals for policy evaluation in adaptive experiments},
  author={Hadad, Vitor and Hirshberg, David A and Zhan, Ruohan and Wager, Stefan and Athey, Susan},
  journal={Proceedings of the national academy of sciences},
  volume={118},
  number={15},
  pages={e2014602118},
  year={2021},
  publisher={National Academy of Sciences}
}

@article{zhang2021statistical,
  title={Statistical inference with m-estimators on adaptively collected data},
  author={Zhang, Kelly and Janson, Lucas and Murphy, Susan},
  journal={Advances in neural information processing systems},
  volume={34},
  pages={7460--7471},
  year={2021}
}

@article{bibaut2025demystifying,
  title={Demystifying inference after adaptive experiments},
  author={Bibaut, Aur{\'e}lien and Kallus, Nathan},
  journal={Annual Review of Statistics and its Application},
  volume={12},
  number={1},
  pages={407--423},
  year={2025},
  publisher={Annual Reviews}
}

@article{hirano2023asymptotic,
  title={Asymptotic representations for sequential decisions, adaptive experiments, and batched bandits},
  author={Hirano, Keisuke and Porter, Jack R},
  journal={arXiv preprint arXiv:2302.03117},
  year={2023}
}

@article{zhang2020inference,
  title={Inference for batched bandits},
  author={Zhang, Kelly and Janson, Lucas and Murphy, Susan},
  journal={Advances in neural information processing systems},
  volume={33},
  pages={9818--9829},
  year={2020}
}

@article{howard2021time_uniform,
  author  = {Howard, Steven R. and Ramdas, Aaditya and McAuliffe, Jon and Sekhon, Jasjeet},
  title   = {Time-Uniform Chernoff Bounds via Nonnegative Supermartingales},
  journal = {Annals of Statistics},
  volume  = {49},
  number  = {2},
  pages   = {1055--1080},
  year    = {2021},
}

@inproceedings{waudby_smith2021tu_clt,
  author    = {Waudby-Smith, Ian and Ramdas, Aaditya},
  title     = {Time-Uniform Central Limit Theorems and Confidence Sequences},
  booktitle = {International Conference on Machine Learning},
  series    = {Proceedings of Machine Learning Research},
  volume    = {139},
  pages     = {10663--10672},
  year      = {2021},
}

@book{LattimoreSzepesvari2020,
  author    = {Lattimore, Tor and Szepesv{\'a}ri, Csaba},
  title     = {Bandit Algorithms},
  publisher = {Cambridge University Press},
  year      = {2020},
  doi       = {10.1017/9781108571401}
}

@inproceedings{GarivierKaufmann2016,
  author    = {Garivier, Aur{\'e}lien and Kaufmann, Emilie},
  title     = {Optimal Best Arm Identification with Fixed Confidence},
  booktitle = {Proceedings of the 29th Conference on Learning Theory (COLT)},
  pages     = {998--1027},
  year      = {2016}
}

@book{ChowChang2011,
  author    = {Chow, Shein-Chung and Chang, Mark},
  title     = {Adaptive Design Methods in Clinical Trials},
  edition   = {2nd},
  publisher = {Chapman \& Hall/CRC},
  year      = {2011}
}

@inproceedings{LiChuLangfordSchapire2010,
  author    = {Li, Lihong and Chu, Wei and Langford, John and Schapire, Robert E.},
  title     = {A Contextual-Bandit Approach to Personalized News Article Recommendation},
  booktitle = {Proceedings of the 19th International Conference on World Wide Web (WWW)},
  pages     = {661--670},
  year      = {2010}
}

@article{QiangBayati2016,
  author    = {Qiang, Sheng and Bayati, Mohsen},
  title     = {Dynamic Pricing with Demand Learning and Strategic Consumers: An Application to Online Retail},
  journal   = {Operations Research},
  volume    = {64},
  number    = {4},
  pages     = {931--944},
  year      = {2016},
  doi       = {10.1287/opre.2016.1514}
}

@article{AtheyEtAl2022,
  author    = {Athey, Susan and Eckles, Dean and Imbens, Guido W.},
  title     = {Design and Analysis of Experiments in the Digital Age},
  journal   = {Annual Review of Economics},
  volume    = {14},
  pages     = {779--806},
  year      = {2022},
  doi       = {10.1146/annurev-economics-051520-023803}
}

@article{CariaEtAl2023,
  author    = {Caria, Stefano and Gordon, Brett and Kasy, Maximilian and others},
  title     = {Adaptive Experiments in Economics},
  journal   = {Annual Review of Economics},
  volume    = {15},
  pages     = {615--647},
  year      = {2023},
  doi       = {10.1146/annurev-economics-091622-031912}
}

@article{perchet2016,
author = {Perchet, Vianney and Rigollet, Philippe and Chassang, Sylvain and Snowberg, Erik},
year = {2016},
month = {04},
pages = {660-681},
title = {Batched Bandit Problems},
volume = {44},
journal = {The Annals of Statistics},
}

@inproceedings{zenati2023sequential,
  title={Sequential counterfactual risk minimization},
  author={Zenati, Houssam and Diemert, Eustache and Martin, Matthieu and Mairal, Julien and Gaillard, Pierre},
  booktitle={International Conference on Machine Learning},
  pages={40681--40706},
  year={2023},
  organization={PMLR}
}

@article{Hill01012011,
author = {Jennifer L. Hill},
title = {Bayesian Nonparametric Modeling for Causal Inference},
journal = {Journal of Computational and Graphical Statistics},
volume = {20},
number = {1},
pages = {217--240},
year = {2011},
publisher = {ASA Website},
doi = {10.1198/jcgs.2010.08162},
URL = { 
https://doi.org/10.1198/jcgs.2010.08162
},
eprint = {        https://doi.org/10.1198/jcgs.2010.08162
}
}

\section*{Checklist}

\begin{enumerate}

  \item For all models and algorithms presented, check if you include:
  \begin{enumerate}
    \item A clear description of the mathematical setting, assumptions, algorithm, and/or model.  
    [Yes] Sections~\ref{section:kte_adaptive}--\ref{section:practical_test_power} define the adaptive data-collection setting, the kernel treatment effect estimand, the projected adaptive test statistic \textsc{ADR-KTE}, and the assumptions used in the analysis. Algorithm~\ref{alg:projected_kte_test} summarizes the practical procedure.
    
    \item An analysis of the properties and complexity (time, space, sample size) of any algorithm.  
    [Partially] The paper provides a statistical analysis of the procedure, including null asymptotic normality, variance-estimation consistency, witness consistency, and consistency under fixed alternatives. It does not provide a dedicated computational complexity analysis in the main text, although implementation details are given in Appendix~\ref{appendix:ClosedFormDerivation}.
    
    \item (Optional) Anonymized source code, with specification of all dependencies, including external libraries.  
    [Yes] Implementation code is provided via the GitHub link in Section~\ref{section:experiments}, and Appendix~\ref{app:additional_experiments} gives additional implementation details.
  \end{enumerate}

  \item For any theoretical claim, check if you include:
  \begin{enumerate}
    \item Statements of the full set of assumptions of all theoretical results.  
    [Yes] See Assumptions~\ref{assum:selection_observables}, \ref{assumption:bounded_kernel}, \ref{assumption:exploration_floor}, \ref{assumption:negligible_clipping}, \ref{assumption:nuisance_rate}, \ref{assumption:policy_complexity}, and \ref{assumption:exploration_floor_strong}.
    
    \item Complete proofs of all theoretical results.  
    [Yes] Proofs are deferred to the appendix, including Appendix~\ref{app:analysis_projected_test}, Appendix~\ref{app:projected_variance_consistency}, and Appendix~\ref{app:witness_power}.
    
    \item Clear explanations of any assumptions.  
    [Yes] The paper provides discussion around the main assumptions, including the role of exploration, the misspecified nuisance limit, the policy-class regularity condition, and the RKHS-specific discussion in Appendix~\ref{app:why_not_rkhs_stabilization}.
  \end{enumerate}

  \item For all figures and tables that present empirical results, check if you include:
  \begin{enumerate}
    \item The code, data, and instructions needed to reproduce the main experimental results (either in the supplemental material or as a URL).  
    [Yes] A code repository is linked in Section~\ref{section:experiments}, and Appendix~\ref{app:additional_experiments} provides additional implementation details.
    
    \item All the training details (e.g., data splits, hyperparameters, how they were chosen).  
    [Yes] Experimental setups, adaptive policies, fold splits, kernels, regularization choices, and data-generation details are described in Section~\ref{section:experiments} and Appendix~\ref{app:additional_experiments}.
    
    \item A clear definition of the specific measure or statistics and error bars (e.g., with respect to the random seed after running experiments multiple times).  
    [Yes] The paper reports rejection rates / true positive rates with standard errors in tables and describes Monte Carlo repetition counts in the experimental sections and appendix. Calibration plots and power plots are also provided.
    
    \item A description of the computing infrastructure used. (e.g., type of GPUs, internal cluster, or cloud provider).  
    [Yes] See the computation-infrastructure paragraph in Appendix~\ref{app:additional_experiments}.
  \end{enumerate}

  \item If you are using existing assets (e.g., code, data, models) or curating/releasing new assets, check if you include:
  \begin{enumerate}
    \item Citations of the creator if your work uses existing assets.  
    [Yes] The paper cites the IHDP dataset \citep{Hill01012011}, dSprites \citep{dsprites17}, and the baseline methods used for comparison.
    
    \item The license information of the assets, if applicable.  
    [No] The current draft cites the datasets and methods but does not explicitly list asset-license information.
    
    \item New assets either in the supplemental material or as a URL, if applicable.  
    [Yes] The implementation is made available through the GitHub repository linked in Section~\ref{section:experiments}.
    
    \item Information about consent from data providers/curators.  
    [Not Applicable]
    
    \item Discussion of sensible content if applicable, e.g., personally identifiable information or offensive content.  
    [Not Applicable] The experiments use standard research benchmark datasets and simulated data, and do not involve sensitive content.
  \end{enumerate}

  \item If you used crowdsourcing or conducted research with human subjects, check if you include:
  \begin{enumerate}
    \item The full text of instructions given to participants and screenshots.  
    [Not Applicable]
    
    \item Descriptions of potential participant risks, with links to Institutional Review Board (IRB) approvals if applicable.  
    [Not Applicable]
    
    \item The estimated hourly wage paid to participants and the total amount spent on participant compensation.  
    [Not Applicable]
  \end{enumerate}

\end{enumerate}

\clearpage
\appendix
\thispagestyle{empty}

\runningtitle{Kernel Treatment Effects with Adaptively Collected Data}

\onecolumn
\aistatstitle{Kernel Treatment Effects with Adaptively Collected Data: Appendix}

This appendix is organized as follows:

\begin{itemize}[nosep, label={--}]
    \item Appendix~\ref{app:notations}: summary of the notations used in the paper and in the analysis.
    \item Appendix~\ref{app:background}: background on reproducing kernel Hilbert spaces and martingale tools used in the proofs.
    \item Appendix~\ref{app:why_not_rkhs_stabilization}: discussion of direct RKHS stabilization and its limitations under adaptive data collection.
    \item Appendix~\ref{app:analysis_projected_test}: proofs for the adaptive projected test statistic introduced in Section~\ref{section:adaptive_test}.
    \item Appendix~\ref{app:projected_variance_consistency}: proof of consistency of the projected conditional standard deviation estimator from Section~\ref{section:variance-estimation}.
    \item Appendix~\ref{app:witness_power}: proofs for asymptotic validity, pilot witness consistency, and consistency under fixed alternatives from Section~\ref{section:practical_test_power}.
    \item Appendix~\ref{appendix:ClosedFormDerivation}: implementation details.
    \item Appendix~\ref{app:additional_experiments}:  additional experimental results.
\end{itemize}

\section{NOTATIONS}
\label{app:notations}

We collect here the notation used in the paper.

\smallskip

\textbf{Adaptive data and filtration}

\smallskip
\begin{itemize}[nosep, label={--}, topsep=-4pt]
    \item $t \in \{1,\ldots,T\}$: round index.
    \item $X_t \in \cX$, $A_t \in \cA$, $Y_t \in \cY$: context, action, and outcome at round $t$.
    \item $\cF_t := \sigma(X_1,A_1,Y_1,\ldots,X_t,A_t,Y_t)$: observed-data filtration.
    \item $X_t \sim P_X$ i.i.d., and $Y_t \sim P_{Y\mid X,A}(\cdot \mid X_t,A_t)$.
    \item $\pi_t(\cdot\mid X_t)$: adaptive logging policy at time $t$, possibly $\cF_{t-1}$-measurable.
    \item $\mu_{\cA}$: base measure on the discrete action space $\cA$.
    \item $\cD_T := \{(X_t,A_t,Y_t)\}_{t=1}^T$: observed trajectory.
\end{itemize}

\smallskip

\textbf{Kernel and embedding notation}

\smallskip
\begin{itemize}[nosep, label={--}, topsep=-4pt]
    \item $k_\cY$: bounded characteristic kernel on $\cY$ with RKHS $\cH_\cY$.
    \item $\phi_\cY(y):=k_\cY(\cdot,y)$: outcome feature map.
    \item $\mu_{Y\mid A,X}(a,x):=\E[\phi_\cY(Y)\mid A=a,X=x]\in\cH_\cY$: conditional mean embedding.
    \item $\eta(a):=\E_{P_X}[\mu_{Y\mid A,X}(a,X)]\in\cH_\cY$: interventional mean embedding.
    \item $\Psi(a,a'):=\eta(a)-\eta(a')\in\cH_\cY$: embedding difference.
    \item $\tau(a,a'):=\|\Psi(a,a')\|_{\cH_\cY}$: kernel treatment effect.
\end{itemize}

\smallskip

\textbf{Doubly robust score}

\smallskip
\begin{itemize}[nosep, label={--}, topsep=-4pt]
    \item For a policy $\pi$ and nuisance $\bar\mu:\cA\times\cX\to\cH_\cY$,
    \[
    D'(\pi,\bar\mu,a)(X,A,Y)
    :=
    \frac{\mathbbm 1\{A=a\}}{\pi(a\mid X)}
    \bigl(\phi_\cY(Y)-\bar\mu(A,X)\bigr)
    + \bar\mu(a,X).
    \]
    \item Foldwise score difference:
    \[
    \hat\phi_t^{(r)}
    :=
    D'(\pi_t,\hat\mu_{t-1}^{(r)},a)(X_t,A_t,Y_t)
    -
    D'(\pi_t,\hat\mu_{t-1}^{(r)},a')(X_t,A_t,Y_t)
    \in \cH_\cY.
    \]
\end{itemize}

\smallskip

\textbf{Sample split and projected statistic}

\smallskip
\begin{itemize}[nosep, label={--}, topsep=-4pt]
    \item We write $T=2n$ and use the chronological split
    \[
    \cI_1=\{1,\ldots,n\},
    \qquad
    \cI_2=\{n+1,\ldots,2n\}.
    \]
    \item Raw pilot witness:
    \[
    v_n
    :=
    \frac{1}{n}\sum_{t\in\cI_1}\hat\phi_t^{(1)}
    \in \cH_\cY.
    \]
    \item Tie-broken pilot direction:
    \[
    w_n
    :=
    \begin{cases}
    v_n/\|v_n\|_{\cH_\cY}, & v_n\neq 0,\\
    u_0, & v_n=0,
    \end{cases}
    \qquad \|u_0\|_{\cH_\cY}=1.
    \]
    \item Projected inferential-fold score:
    \[
    \xi_t(w_n):=\langle w_n,\hat\phi_t^{(2)}\rangle_{\cH_\cY},
    \qquad t\in\cI_2.
    \]
    \item Projected target:
    \[
    \theta_n(a,a'):=\langle w_n,\Psi(a,a')\rangle_{\cH_\cY}.
    \]
\end{itemize}

\smallskip

\textbf{Sequential conditional variance estimation}

\smallskip
\begin{itemize}[nosep, label={--}, topsep=-4pt]
    \item For $t\in\cI_2$, the oracle projected conditional standard deviation is
    \[
    \sigma_{0,t}(w_n):=
    \Var(\xi_t(w_n)\mid \cF_{t-1})^{1/2}.
    \]
    \item Past index set in the inferential fold:
    \[
    S_t:=\{s\in\cI_2:s<t\},
    \qquad
    m_t:=|S_t|.
    \]
    \item History-recycled RKHS score:
    \[
    \check\phi_{s,t}^{(2)}
    :=
    D'(\pi_t,\hat\mu_{s-1}^{(2)},a)(X_s,A_s,Y_s)
    -
    D'(\pi_t,\hat\mu_{s-1}^{(2)},a')(X_s,A_s,Y_s).
    \]
    \item Importance ratio:
    \[
    w_{s,t}:=\frac{\pi_t(A_s\mid X_s)}{\pi_s(A_s\mid X_s)}.
    \]
    \item History-recycled projected score:
    \[
    \check\xi_{s,t}(w_n):=
    \langle w_n,\check\phi_{s,t}^{(2)}\rangle_{\cH_\cY}.
    \]
    \item Plug-in projected moments for $t\ge n+2$:
    \[
    \hat M_{1,t}(w_n):=
    \frac{1}{m_t}\sum_{s\in S_t} w_{s,t}\check\xi_{s,t}(w_n),
    \]
    \[
    \hat M_{2,t}(w_n):=
    \frac{1}{m_t}\sum_{s\in S_t} w_{s,t}\check\xi_{s,t}(w_n)^2.
    \]
    \item Initial convention at the first inferential time:
    \[
    \hat M_{1,n+1}(w_n):=0,
    \qquad
    \hat M_{2,n+1}(w_n):=1,
    \qquad
    \hat\sigma_{n+1}^2(w_n):=1.
    \]
    \item Plug-in conditional variance and standard deviation for $t\ge n+2$:
    \[
    \hat\sigma_t^2(w_n):=
    \bigl(\hat M_{2,t}(w_n)-\hat M_{1,t}(w_n)^2\bigr)_+,
    \qquad
    \hat\sigma_t(w_n):=\sqrt{\hat\sigma_t^2(w_n)}.
    \]
    \item Clipped feasible standard deviation:
    \[
    \tilde\sigma_t(w_n):=\hat\sigma_t(w_n)\vee \epsilon_n.
    \]
\end{itemize}

\smallskip

\textbf{Final adaptive test statistic}

\smallskip
\begin{itemize}[nosep, label={--}, topsep=-4pt]
    \item Final adaptive test statistic:
    \[
    T_n(w_n):=
    \frac{1}{\sqrt n}
    \sum_{t\in\cI_2}\tilde\sigma_t(w_n)^{-1}\xi_t(w_n).
    \]
\end{itemize}

\smallskip

\textbf{Reference policy class used in Appendix~\ref{app:projected_variance_consistency}}

\smallskip
\begin{itemize}[nosep, label={--}, topsep=-4pt]
    \item $\pi_{\mathrm{ref}}$: fixed strictly positive reference density.
    \item $\mathcal G_{\mathrm{ref}}:=\{\pi/\pi_{\mathrm{ref}}:\pi\in\Pi\}$.
    \item
    \[
    \|h\|_{2,\mathrm{ref}}^2
    :=
    \E_{X\sim P_X}\!\left[
      \int h(b,X)^2\,\pi_{\mathrm{ref}}(b\mid X)\,\diff\mu_{\cA}(b)
    \right].
    \]
\end{itemize}

\section{BACKGROUND}
\label{app:background}

This appendix collects the background material used in the current proof stack. 
\subsection{Review of Reproducing Kernel Hilbert Spaces}
\label{app:rkhs_review}

A \emph{positive definite kernel} on a set \(\mathcal F\) is a function
\(k:\mathcal F\times\mathcal F\to\mathbb R\) such that for any
\(m\in\mathbb N\), any \(w_1,\dots,w_m\in\mathcal F\), and any
\(c_1,\dots,c_m\in\mathbb R\),
\[
\sum_{i,j=1}^m c_i c_j\,k(w_i,w_j)\ge 0.
\]
By the Moore--Aronszajn theorem, \(k\) induces a unique Hilbert space
\(\mathcal H_{\mathcal F}\) of functions on \(\mathcal F\) with inner product
\(\langle\cdot,\cdot\rangle_{\mathcal H_{\mathcal F}}\) such that
\(k(\cdot,w)\in\mathcal H_{\mathcal F}\) for every \(w\in\mathcal F\), and the reproducing property holds:
\[
f(w)=\langle f,k(\cdot,w)\rangle_{\mathcal H_{\mathcal F}}
\qquad
\forall f\in\mathcal H_{\mathcal F},\ \forall w\in\mathcal F.
\]
We write the associated feature map as
\[
\phi_{\mathcal F}(w):=k(\cdot,w)\in\mathcal H_{\mathcal F}.
\]

\paragraph{Kernel mean embeddings.}
Let \(W\sim P\) be a random element of \(\mathcal F\) such that
\(\E[\sqrt{k(W,W)}]<\infty\). The \emph{kernel mean embedding} of \(P\) into
\(\mathcal H_{\mathcal F}\) is
\[
\mu_P:=\E[\phi_{\mathcal F}(W)]\in\mathcal H_{\mathcal F}.
\]
Given observations \(w_1,\dots,w_n\), the empirical embedding is
\[
\widehat\mu_P=\frac1n\sum_{i=1}^n \phi_{\mathcal F}(w_i).
\]

\paragraph{Conditional mean embeddings.}
Let \(X\in\mathcal X\) and \(Y\in\mathcal Y\) be random variables with RKHSs
\(\mathcal H_{\mathcal X}\) and \(\mathcal H_{\mathcal Y}\), and associated
feature maps \(\phi_{\mathcal X}\) and \(\phi_{\mathcal Y}\). Define the covariance operators
\[
C_{YX}:=\E[\phi_{\mathcal Y}(Y)\otimes \phi_{\mathcal X}(X)],
\qquad
C_{XX}:=\E[\phi_{\mathcal X}(X)\otimes \phi_{\mathcal X}(X)].
\]
When \(C_{XX}\) is injective, the conditional mean operator is
\[
\mathcal C_{Y\mid X}:=C_{YX}C_{XX}^{-1},
\]
and the conditional mean embedding satisfies
\[
\mu_{Y\mid X}(x)
=
\mathcal C_{Y\mid X}\phi_{\mathcal X}(x)
=
\E[\phi_{\mathcal Y}(Y)\mid X=x].
\]
With data \(\{(x_i,y_i)\}_{i=1}^n\), a standard regularized estimator is
\[
\widehat{\mathcal C}_{Y\mid X}
=
\Phi_Y (K_X+\lambda I_n)^{-1}\Phi_X^\top,
\qquad
\widehat\mu_{Y\mid X}(x)
=
\widehat{\mathcal C}_{Y\mid X}\phi_{\mathcal X}(x),
\]
where \(K_X\) is the Gram matrix on \(\{x_i\}_{i=1}^n\), and
\(\Phi_X,\Phi_Y\) collect the feature maps in their columns; see
\citet{song2009hilbert,li2022optimal}.

\paragraph{Maximum mean discrepancy.}
For two distributions \(P,Q\) on \(\mathcal F\), the maximum mean discrepancy is
\[
\mathrm{MMD}(P,Q):=\|\mu_P-\mu_Q\|_{\mathcal H_{\mathcal F}}.
\]
If \(k\) is characteristic, then \(\mathrm{MMD}(P,Q)=0\) if and only if
\(P=Q\) \citep{gretton2012kernel}. Under i.i.d.\ sampling, direct quadratic kernel statistics are degenerate under the null, and recent cross-\(U\) constructions restore asymptotic normality by sample splitting \citep{kim2024dimension}. Our procedure adopts the same broad philosophy, but under adaptive data collection and after projection onto a pilot witness.

\paragraph{Interventional mean embeddings and KTE.}
In our setting, \(\mathcal F=\mathcal Y\) and the corresponding RKHS is
\(\mathcal H_{\mathcal Y}\). The interventional mean embedding under action
\(a\in\mathcal A\) is
\[
\eta(a):=\E_{P_X}[\mu_{Y\mid A,X}(a,X)]\in\mathcal H_{\mathcal Y},
\]
and the kernel treatment effect between actions \(a\) and \(a'\) is
\[
\tau(a,a')=\|\eta(a)-\eta(a')\|_{\mathcal H_{\mathcal Y}}.
\]

\subsection{Martingale Tools}

Let \((\mathcal F_t)_{t\ge 0}\) be a filtration. A sequence
\((Z_t)_{t\ge 1}\) of \(\mathcal H\)-valued random elements is a martingale
difference sequence if \(Z_t\) is \(\mathcal F_t\)-measurable and
\[
\E[Z_t\mid \mathcal F_{t-1}]=0.
\]

\paragraph{Martingale orthogonality.}
If \((Z_t)\) is square-integrable and \(\mathcal H\)-valued, then for
\(s\neq t\),
\begin{equation}
\label{eq:MO}
\E\langle Z_s,Z_t\rangle_{\mathcal H}=0.
\end{equation}
Indeed, if \(s<t\), then \(Z_s\) is \(\mathcal F_{t-1}\)-measurable, so
\[
\E\langle Z_s,Z_t\rangle
=
\E\!\left[
\left\langle
Z_s,\E[Z_t\mid \mathcal F_{t-1}]
\right\rangle
\right]
=0.
\]

As a consequence, if \((Z_t)\) is square-integrable, then
\[
\E\Big\|\sum_{t=1}^n Z_t\Big\|_{\mathcal H}^2
=
\sum_{t=1}^n \E\|Z_t\|_{\mathcal H}^2.
\]

\begin{thm}[Strong law for martingale sums {\citep[Thm.~2.18]{hall1980}}]
\label{thm:HH-SLLN}
Let \(\{S_n=\sum_{i=1}^n X_i,\mathcal F_n,n\ge 1\}\) be a scalar martingale and let \(\{U_n,n\ge 1\}\) be a nondecreasing sequence of positive random variables such that \(U_n\) is \(\mathcal F_{n-1}\)-measurable for each \(n\). If \(1\le p\le 2\), then
\[
\lim_{n\to\infty} U_n^{-1}S_n = 0
\]
almost surely on the set
\[
\left\{
\lim_{n\to\infty} U_n=\infty,\ 
\sum_{i=1}^\infty U_i^{-p}\E(|X_i|^p\mid\mathcal F_{i-1})<\infty
\right\}.
\]
\end{thm}

\begin{rem}[How we use Theorem~\ref{thm:HH-SLLN}]
Taking \(p=2\) and \(U_n=n\) yields
\[
\frac1n\sum_{i=1}^n X_i \xrightarrow{\text{a.s.}} 0
\]
whenever
\[
\sum_{i=1}^\infty i^{-2}\E[X_i^2\mid \mathcal F_{i-1}]<\infty.
\]
This is the form used to control centered scalar remainder terms in the variance-estimation argument.
\end{rem}

\begin{thm}[Scalar martingale CLT with predictable-variation normalization]
\label{thm:scalar_martingale_clt}
Let \((X_t,\mathcal F_t)_{t\ge 1}\) be a square-integrable scalar martingale difference sequence and define
\[
V_n^2
:=
\sum_{t=1}^n \E[X_t^2\mid \mathcal F_{t-1}].
\]
Assume \(V_n^2\to\infty\) in probability and
\[
\frac{\max_{1\le t\le n}|X_t|}{V_n}\xrightarrow[]{p}0.
\]
Then
\[
\frac{\sum_{t=1}^n X_t}{V_n}
\ \xRightarrow[]{d}\
\mathcal N(0,1).
\]
\end{thm}

\begin{thm}[Scalar martingale CLT for triangular arrays]
\label{thm:scalar_array_clt}
Let \(\{(X_{n,t},\mathcal F_{n,t}) : 1\le t\le n,\ n\ge1\}\) be a square-integrable scalar martingale-difference array. Assume
\[
\sum_{t=1}^n \E[X_{n,t}^2\mid \mathcal F_{n,t-1}]
\xrightarrow[]{p} 1
\]
and that there exists \(\delta>0\) such that
\[
\sum_{t=1}^n \E[|X_{n,t}|^{2+\delta}\mid \mathcal F_{n,t-1}]
\xrightarrow[]{p} 0.
\]
Then
\[
\sum_{t=1}^n X_{n,t}
\ \xRightarrow[]{d}\
\mathcal N(0,1).
\]
\end{thm}

\section{DIRECT RKHS STABILIZATION AND ITS LIMITATIONS}
\label{app:why_not_rkhs_stabilization}

A natural route for adaptive inference with kernel treatment effects is to work directly with the RKHS-valued score process
\[
\hat\phi_t
=
D'(\pi_t,\widehat\mu_{t-1},a)(X_t,A_t,Y_t)
-
D'(\pi_t,\widehat\mu_{t-1},a')(X_t,A_t,Y_t)
\in \cH_{\cY},
\]
and to study asymptotic normality of a normalized RKHS-valued average. This appendix discusses the scope and limitations of that approach.

\subsection{Uniform Positivity and Bounded RKHS Scores}

Under uniform positivity, the raw RKHS score sequence is already uniformly bounded. In that regime, one can study the unscaled RKHS average directly through standard martingale arguments.

\begin{prop}[Uniform positivity yields bounded RKHS scores]
\label{prop:bounded_raw_scores}
Assume:
\begin{enumerate}[label=(\roman*), nosep]
    \item Assumption~\ref{assum:selection_observables},
    \item the bounded-kernel part of Assumption~\ref{assumption:bounded_kernel},
    \item the nuisance envelope
    \begin{equation}
    \label{eq:appendix_nuisance_envelope}
    \sup_{t\ge1}\sup_{a\in\cA,\ x\in\cX}
    \|\widehat\mu_{t-1}(a,x)\|_{\cH_{\cY}}
    \le B_\mu
    \quad \text{a.s.},
    \end{equation}
    \item a uniform positivity condition: there exists \(c>0\) such that
    \[
    \pi_t(a\mid x)\ge c
    \qquad
    \forall t\ge1,\ \forall a\in\cA,\ \text{for }P_X\text{-a.e. }x.
    \]
\end{enumerate}
Then there exists \(C<\infty\) such that
\[
\|\hat\phi_t\|_{\cH_{\cY}}\le C
\qquad \text{a.s. for all } t\ge1.
\]
\end{prop}

\begin{proof}
By boundedness of the kernel,
\[
\|\phi_{\cY}(Y_t)\|_{\cH_{\cY}}\le \sqrt{\kappa}.
\]
Hence for any \(b\in\cA\),
\[
\big\|
D'(\pi_t,\widehat\mu_{t-1},b)(X_t,A_t,Y_t)
\big\|_{\cH_{\cY}}
\le
\frac{1}{c}\big(\sqrt{\kappa}+B_\mu\big)+B_\mu.
\]
The claim follows by the triangle inequality applied to the difference between the two actions \(a\) and \(a'\).
\end{proof}

Proposition~\ref{prop:bounded_raw_scores} shows that, under uniform positivity, the score process is already uniformly controlled in norm. In such settings, variance normalization is not needed to prevent score increments from becoming large.

\subsection{Trace Normalization and Operator Geometry}

Let
\[
\Sigma_t:=\Cov(\hat\phi_t\mid \cF_{t-1})
\]
denote the conditional covariance operator of the RKHS score. A natural scalar normalization is obtained from its trace:
\[
\omega_t^{-2}:=\Tr(\Sigma_t),
\qquad
\widetilde\Sigma_t
:=
\omega_t^2\Sigma_t
=
\frac{\Sigma_t}{\Tr(\Sigma_t)}.
\]
This normalization fixes the total conditional variance:
\[
\Tr(\widetilde\Sigma_t)=1.
\]
However, it does not determine the directional geometry of the covariance operator in \(\cH_{\cY}\). Even after normalization by trace, the operators \(\widetilde\Sigma_t\) may continue to vary across different directions of the RKHS.

For \(u,v\in\cH_{\cY}\), recall that the rank-one operator \(u\otimes v\) is defined by
\[
(u\otimes v)h:=\langle v,h\rangle_{\cH_{\cY}}\,u,
\qquad h\in\cH_{\cY}.
\]
This notation is sufficient to make the issue explicit.

\subsection{A Finite-Rank Illustration}

\begin{exmp}[Directional variation after trace normalization]
\label{exmp:directional_drift}
Let \(u_1,u_2\in \cH_{\cY}\) be orthonormal, let \((\lambda_t)_{t\ge1}\) be any positive scalar sequence, and define
\[
\Sigma_t
=
\lambda_t\,u_1\otimes u_1
\quad \text{for } t\in B_{2m-1},
\qquad
\Sigma_t
=
\lambda_t\,u_2\otimes u_2
\quad \text{for } t\in B_{2m},
\]
where the block lengths satisfy \(|B_m|=2^{2^m}\). Then
\[
\omega_t^{-2}=\Tr(\Sigma_t)=\lambda_t,
\qquad
\widetilde\Sigma_t
=
\frac{\Sigma_t}{\Tr(\Sigma_t)}
=
\begin{cases}
u_1\otimes u_1, & t\in B_{2m-1},\\
u_2\otimes u_2, & t\in B_{2m}.
\end{cases}
\]
Thus trace normalization removes the scalar factor \(\lambda_t\), but the normalized covariance still alternates between two distinct directions in \(\cH_{\cY}\). Because the block lengths grow super-exponentially, the Cesàro averages of \((\widetilde\Sigma_t)\) along the terminal times of odd and even blocks have different limits. In particular, \((\widetilde\Sigma_t)\) has no Cesàro limit.
\end{exmp}

Example~\ref{exmp:directional_drift} shows that trace normalization controls the scale of the covariance operator but not its directional variation. Consequently, scalar normalization by \(\Tr(\Sigma_t)\) alone is not sufficient to guarantee stabilization of an RKHS-valued covariance process.

\begin{rem}[Why we do not use a CADR-type nondegenerate efficiency bound in the RKHS]
\label{rem:rkhs_nondegeneracy_false}
The scalar nondegeneracy condition used in CADR \citep{bibaut2021post} is natural in a one-dimensional problem. Its literal RKHS analogue would require a lower bound of the form
\[
\Sigma_t \succeq cI
\]
for the conditional covariance operator of the RKHS score. In an infinite-dimensional RKHS, conditional covariance operators are typically compact, so such a lower bound fails on any infinite-dimensional subspace. This is why the main null theorem is formulated through direct control of the feasible predictable variation of the projected statistic rather than through a uniform operator-level nondegeneracy assumption.
\end{rem}

\subsection{Projection and Scalar Stabilization}

The preceding discussion motivates reducing the inferential problem before normalization. Once the score is projected onto a fixed direction \(v_n\in \cH_{\cY}\),
\[
\xi_t(v_n)=\langle v_n,\hat\phi_t\rangle_{\cH_{\cY}},
\]
the resulting sequence is scalar. At that point, normalization by a conditional standard deviation becomes a direct stabilization of the object entering the final test.

This projection step preserves the distributional nature of the problem through the choice of \(v_n\), while avoiding the need to stabilize an operator-valued process in the full RKHS. The resulting inference problem is therefore governed by the scalar sequence \((\xi_t(v_n))\) rather than by the full RKHS score process \((\hat\phi_t)\).

The discussion above does not rule out RKHS-valued stabilization in principle. It shows, rather, that trace normalization alone does not control the operator-level variation needed for RKHS-valued Gaussian limits under general adaptive sampling, whereas projection reduces the normalization problem to a scalar one.

\section{PROOFS FOR THE ADAPTIVE TEST}
\label{app:analysis_projected_test}

This appendix contains the proofs for the results stated in Section~\ref{section:adaptive_test}. We first prove the projected doubly robust identity. We then establish projected moment bounds, verify the predictable variation and Lyapunov conditions of the feasible statistic, and finally prove null asymptotic normality.

\subsection{Projected doubly robust identity}

\begin{proof}[Proof of Lemma~\ref{lem:projected_dr_identity}]
Since \(w_n\) is \(\cF_n\)-measurable and \(n<t\), it is \(\cF_{t-1}\)-measurable. Therefore
\[
\E\!\left[
\xi_t(w_n)
\,\middle|\,
\cF_{t-1}
\right]
=
\left\langle
w_n,\,
\E\!\left[\hat\phi_t^{(2)}\mid \cF_{t-1}\right]
\right\rangle_{\cH_{\cY}}.
\]
It suffices to show that
\[
\E\!\left[\hat\phi_t^{(2)}\mid \cF_{t-1}\right]
=
\Psi(a,a').
\]
Fix \(b\in\{a,a'\}\). For any predictable nuisance \(\bar\mu\),
\[
\E\!\left[
D'(\pi_t,\bar\mu,b)(X_t,A_t,Y_t)
\,\middle|\,
\cF_{t-1}
\right]
=
\eta(b),
\]
by the standard doubly robust identity. Taking the difference between \(b=a\) and \(b=a'\) yields
\[
\E\!\left[\hat\phi_t^{(2)}\mid \cF_{t-1}\right]
=
\eta(a)-\eta(a')
=
\Psi(a,a').
\]
Hence
\[
\E\!\left[
\xi_t(w_n)
\,\middle|\,
\cF_{t-1}
\right]
=
\langle w_n,\Psi(a,a')\rangle_{\cH_{\cY}}
=:
\theta_n(a,a').
\]
Under \(H_0:\eta(a)=\eta(a')\), this equals \(0\).
\end{proof}

\subsection{Projected moment bounds}

\begin{lem}[Projected \(q\)-th moment bound]
\label{lem:projected_qth_moment_bound}
Assume Assumptions~\ref{assum:selection_observables},
\ref{assumption:bounded_kernel},
\eqref{eq:nuisance_envelope}, and
\ref{assumption:exploration_floor}. Then for every \(q\ge 2\) there exists \(C_q<\infty\) such that, for every \(t\in\mathcal I_2\),
\[
\E\!\left[
|\xi_t(w_n)|^q
\,\middle|\,
\cF_{t-1}
\right]
\le
C_q\, t^{\alpha(q-1)}
\qquad\text{a.s.}
\]
In particular,
\[
\E\!\left[
\xi_t(w_n)^2
\,\middle|\,
\cF_{t-1}
\right]
\lesssim t^\alpha,
\qquad
\E\!\left[
\xi_t(w_n)^4
\,\middle|\,
\cF_{t-1}
\right]
\lesssim t^{3\alpha},
\]
and therefore
\[
\sigma_{0,t}^2(w_n)\lesssim t^\alpha
\qquad\text{a.s.}
\]
\end{lem}

\begin{proof}
Since \(\|w_n\|_{\cH_{\cY}}=1\),
\[
|\xi_t(w_n)|^q
=
|\langle w_n,\hat\phi_t^{(2)}\rangle_{\cH_{\cY}}|^q
\le
\|\hat\phi_t^{(2)}\|_{\cH_{\cY}}^q.
\]
Write
\[
\hat\phi_t^{(2)} = G_t(a)-G_t(a'),
\qquad
G_t(b):=D'(\pi_t,\widehat\mu_{t-1}^{(2)},b)(X_t,A_t,Y_t).
\]
By \((u+v)^q\le 2^{q-1}(u^q+v^q)\),
\[
\|\hat\phi_t^{(2)}\|_{\cH_{\cY}}^q
\le
2^{q-1}\|G_t(a)\|_{\cH_{\cY}}^q
+
2^{q-1}\|G_t(a')\|_{\cH_{\cY}}^q.
\]
Fix \(b\in\{a,a'\}\). By boundedness of the kernel and the nuisance envelope,
\[
\big\|
\phi_\cY(Y_t)-\widehat\mu_{t-1}^{(2)}(A_t,X_t)
\big\|_{\cH_{\cY}}
\le
\sqrt\kappa+B_\mu
=:C_\mu,
\]
and
\[
\|\widehat\mu_{t-1}^{(2)}(b,X_t)\|_{\cH_{\cY}}
\le
B_\mu.
\]
Therefore
\[
\|G_t(b)\|_{\cH_{\cY}}
\le
C_\mu \frac{\mathbbm 1\{A_t=b\}}{\pi_t(b\mid X_t)} + B_\mu,
\]
so
\[
\|G_t(b)\|_{\cH_{\cY}}^q
\lesssim
\frac{\mathbbm 1\{A_t=b\}}{\pi_t(b\mid X_t)^q}
+
1.
\]
Taking conditional expectations and using
\[
\E\!\left[
\frac{\mathbbm 1\{A_t=b\}}{\pi_t(b\mid X_t)^q}
\,\middle|\,
\cF_{t-1}
\right]
=
\E_{X_t}
\!\left[
\pi_t(b\mid X_t)^{-(q-1)}
\right]
\lesssim
t^{\alpha(q-1)}
\]
by Assumption~\ref{assumption:exploration_floor}, we obtain
\[
\E\!\left[
\|G_t(b)\|_{\cH_{\cY}}^q
\,\middle|\,
\cF_{t-1}
\right]
\lesssim
t^{\alpha(q-1)}.
\]
The same bound therefore holds for \(\|\hat\phi_t^{(2)}\|_{\cH_{\cY}}^q\) and hence for \(|\xi_t(w_n)|^q\). The variance bound follows from
\[
\sigma_{0,t}^2(w_n)
\le
\E\!\left[\xi_t(w_n)^2\mid \cF_{t-1}\right].
\qedhere
\]
\end{proof}

\subsection{Predictable variation of the feasible statistic}

Define
\[
X_{n,t}
:=
\frac{1}{\sqrt n}\,
\widetilde\sigma_t(w_n)^{-1}\,\xi_t(w_n),
\qquad t\in\mathcal I_2.
\]
Under \(H_0\), Lemma~\ref{lem:projected_dr_identity} shows that \((X_{n,t},\cF_t)_{t\in\mathcal I_2}\) is a scalar martingale-difference array.

\begin{lem}[Predictable variation convergence]
\label{lem:predictable_variation_feasible}
Assume the conditions of Theorem~\ref{thm:projected_test_clt}. Then
\[
V_n^2
:=
\sum_{t\in\mathcal I_2}
\E\!\left[
X_{n,t}^2
\,\middle|\,
\cF_{t-1}
\right]
=
\frac{1}{n}
\sum_{t\in\mathcal I_2}
\frac{\sigma_{0,t}^2(w_n)}{\widetilde\sigma_t^2(w_n)}
\xrightarrow[]{p}1.
\]
\end{lem}

\begin{proof}
This is exactly \eqref{eq:predictable_variation_consistency} from Corollary~\ref{assumption:average_inverse_sd_consistency}.
\end{proof}

\subsection{Lyapunov condition for the feasible statistic}

\begin{lem}[Feasible Lyapunov condition]
\label{lem:feasible_lyapunov}
Assume the conditions of Theorem~\ref{thm:projected_test_clt}. Then
\[
\sum_{t\in\mathcal I_2}
\E\!\left[
|X_{n,t}|^4
\,\middle|\,
\cF_{t-1}
\right]
\xrightarrow[]{p}0.
\]
\end{lem}

\begin{proof}
By definition of \(X_{n,t}\),
\[
\sum_{t\in\mathcal I_2}
\E\!\left[
|X_{n,t}|^4
\,\middle|\,
\cF_{t-1}
\right]
=
\frac{1}{n^2}
\sum_{t\in\mathcal I_2}
\widetilde\sigma_t(w_n)^{-4}
\E\!\left[
\xi_t(w_n)^4
\,\middle|\,
\cF_{t-1}
\right].
\]
Since \(\widetilde\sigma_t(w_n)\ge \epsilon_n=n^{-\rho}\), Lemma~\ref{lem:projected_qth_moment_bound} yields
\[
\sum_{t\in\mathcal I_2}
\E\!\left[
|X_{n,t}|^4
\,\middle|\,
\cF_{t-1}
\right]
\lesssim
n^{-2}\epsilon_n^{-4}
\sum_{t=n+1}^{2n} t^{3\alpha}
\lesssim
n^{-2+4\rho} \cdot n^{1+3\alpha}
=
n^{3\alpha-1+4\rho}.
\]
By \eqref{eq:clip_rate_choice}, \(\rho<(1-3\alpha)/4\), so the right-hand side converges to \(0\).
\end{proof}

\subsection{Proof of the null theorem}

\begin{proof}[Proof of Theorem~\ref{thm:projected_test_clt}]
Under \(H_0\), Lemma~\ref{lem:projected_dr_identity} implies that \((X_{n,t},\cF_t)_{t\in\mathcal I_2}\) is a scalar martingale-difference array. By Lemma~\ref{lem:predictable_variation_feasible},
\[
\sum_{t\in\mathcal I_2}
\E\!\left[
X_{n,t}^2
\,\middle|\,
\cF_{t-1}
\right]
\xrightarrow[]{p}1.
\]
By Lemma~\ref{lem:feasible_lyapunov},
\[
\sum_{t\in\mathcal I_2}
\E\!\left[
|X_{n,t}|^4
\,\middle|\,
\cF_{t-1}
\right]
\xrightarrow[]{p}0.
\]
Applying the scalar martingale CLT for triangular arrays, Theorem~\ref{thm:scalar_array_clt}, gives
\[
\sum_{t\in\mathcal I_2} X_{n,t}
\ \xRightarrow[]{d}\
\mathcal N(0,1).
\]
By definition,
\[
\sum_{t\in\mathcal I_2} X_{n,t}
=
\frac{1}{\sqrt n}
\sum_{t\in\mathcal I_2}
\widetilde\sigma_t(w_n)^{-1}\xi_t(w_n)
=
T_n(w_n).
\]
This proves the claim.
\end{proof}

\section{PROOF OF THE PROJECTED CONDITIONAL STANDARD DEVIATION CONSISTENCY THEOREM}
\label{app:projected_variance_consistency}

This appendix proves Theorem~\ref{thm:projected_variance_consistency}. The proof does not use RKHS empirical-process argument. Conditionally on the pilot fold, the direction \(w_n\) is fixed and the inferential-fold problem becomes scalar \citep{bibaut2021post}.

Throughout this appendix, condition on \(\cF_n\). By construction,
\begin{equation}
\label{eq:appendix_wn_bound}
\|w_n\|_{\cH_{\cY}}=1.
\end{equation}
All almost-sure statements below are understood conditionally on \(\cF_n\).

\subsection{Scalarization of the projected inferential-fold problem}

For \(s\in\mathcal I_2\), define the scalar projected outcome
\[
\widetilde Y_s
:=
\langle w_n,\phi_\cY(Y_s)\rangle_{\cH_{\cY}},
\]
and the projected nuisance sequence
\[
\widetilde\mu_{s-1}(b,x)
:=
\left\langle
w_n,\widehat\mu^{(2)}_{s-1}(b,x)
\right\rangle_{\cH_{\cY}},
\qquad b\in\cA,\ x\in\cX.
\]
Also define the projected nuisance contrast
\[
\Delta\widetilde\mu_{s-1}(x)
:=
\widetilde\mu_{s-1}(a,x)-\widetilde\mu_{s-1}(a',x).
\]

Since \(\|w_n\|_{\cH_{\cY}}=1\), boundedness of the kernel and the nuisance envelope imply
\begin{equation}
\label{eq:scalarized_envelope}
|\widetilde Y_s|
\le \sqrt\kappa,
\qquad
\sup_{s\ge1}\sup_{b\in\cA,\ x\in\cX}
|\widetilde\mu_{s-1}(b,x)|
\le B_\mu
\quad\text{a.s.}
\end{equation}
Likewise, Assumption~\ref{assumption:nuisance_rate} implies
\begin{equation}
\label{eq:scalarized_nuisance_rate}
\big\|
\widetilde\mu_{t-1}-\widetilde\mu_{\infty}
\big\|_{L_2(P_X\times\mu_{\cA})}
=
O(t^{-\beta})
\qquad\text{a.s.},
\end{equation}
where
\[
\widetilde\mu_\infty(b,x)
:=
\langle w_n,\mu_\infty(b,x)\rangle_{\cH_{\cY}}.
\]

For \(s<t\) in \(\mathcal I_2\), the history-recycled projected score can be written as
\begin{equation}
\label{eq:scalarized_check_xi}
\begin{aligned}
\check\xi_{s,t}(w_n)
={}&
\frac{\mathbbm 1\{A_s=a\}}{\pi_t(a\mid X_s)}
\Big(
\widetilde Y_s-\widetilde\mu_{s-1}(a,X_s)
\Big)
\\
&\quad-
\frac{\mathbbm 1\{A_s=a'\}}{\pi_t(a'\mid X_s)}
\Big(
\widetilde Y_s-\widetilde\mu_{s-1}(a',X_s)
\Big)
\\
&\quad+
\Delta\widetilde\mu_{s-1}(X_s).
\end{aligned}
\end{equation}
Thus, conditionally on \(\cF_n\), the projected inferential-fold score is exactly a scalar doubly robust contrast between the two target actions \(a\) and \(a'\).

\subsection{Projected \(f_1,\dots,f_5\) decomposition}

Let \(g_t:=\pi_t/\pi_{\mathrm{ref}}\). For any \(g\in\mathcal G_{\mathrm{ref}}\) and any scalar nuisance \(m:\cA\times\cX\to\mathbb R\), define
\[
\Delta m(x):=m(a,x)-m(a',x),
\]
\begin{align}
\label{eq:projected_f1}
f_1(g,m)(X,A,Y)
&:=
\frac{\mathbbm 1\{A=a\}}{g(a\mid X)\pi_{\mathrm{ref}}(a\mid X)^2}
\Big(
\widetilde Y-m(a,X)
\Big)^2
\nonumber\\
&\quad+
\frac{\mathbbm 1\{A=a'\}}{g(a'\mid X)\pi_{\mathrm{ref}}(a'\mid X)^2}
\Big(
\widetilde Y-m(a',X)
\Big)^2,
\\
\label{eq:projected_f2}
f_2(m)(X,A,Y)
&:=
2\Bigg[
\frac{\mathbbm 1\{A=a\}}{\pi_{\mathrm{ref}}(a\mid X)}
\Big(
\widetilde Y-m(a,X)
\Big)
\nonumber\\
&\qquad\qquad-
\frac{\mathbbm 1\{A=a'\}}{\pi_{\mathrm{ref}}(a'\mid X)}
\Big(
\widetilde Y-m(a',X)
\Big)
\Bigg]
\Delta m(X),
\\
\label{eq:projected_f3}
f_3(g,m)(X,A,Y)
&:=
g(A\mid X)\,\Delta m(X)^2,
\\
\label{eq:projected_f4}
f_4(m)(X,A,Y)
&:=
\frac{\mathbbm 1\{A=a\}}{\pi_{\mathrm{ref}}(a\mid X)}
\Big(
\widetilde Y-m(a,X)
\Big)
\nonumber\\
&\quad-
\frac{\mathbbm 1\{A=a'\}}{\pi_{\mathrm{ref}}(a'\mid X)}
\Big(
\widetilde Y-m(a',X)
\Big),
\\
\label{eq:projected_f5}
f_5(g,m)(X,A,Y)
&:=
g(A\mid X)\,\Delta m(X).
\end{align}

\begin{lem}[Projected weighted-moment decomposition]
\label{lem:projected_f_decomposition}
For every \(s<t\) in \(\mathcal I_2\),
\begin{align}
\label{eq:projected_second_decomposition}
g_t(A_s\mid X_s)\,
\check\xi_{s,t}(w_n)^2
&=
f_1(g_t,\widetilde\mu_{s-1})(O_s)
+
f_2(\widetilde\mu_{s-1})(O_s)
+
f_3(g_t,\widetilde\mu_{s-1})(O_s),
\\
\label{eq:projected_first_decomposition}
g_t(A_s\mid X_s)\,
\check\xi_{s,t}(w_n)
&=
f_4(\widetilde\mu_{s-1})(O_s)
+
f_5(g_t,\widetilde\mu_{s-1})(O_s).
\end{align}
\end{lem}

\begin{proof}
This is a direct expansion of \eqref{eq:scalarized_check_xi}. The mixed product between the \(a\)- and \(a'\)-terms vanishes because the indicators \(\mathbbm 1\{A_s=a\}\) and \(\mathbbm 1\{A_s=a'\}\) are disjoint.
\end{proof}

\subsection{Empirical and target moment pieces}

For \(t\in\mathcal I_2\) and \(t\ge n+2\), define
\[
\widehat\Phi_{1,t}^{(k)}
:=
\frac{1}{m_t}
\sum_{s\in S_t}
\frac{\pi_{\mathrm{ref}}(A_s\mid X_s)}{\pi_s(A_s\mid X_s)}
\,f_k(g_t,\widetilde\mu_{s-1})(O_s),
\qquad k\in\{1,3,5\},
\]
\[
\widehat\Phi_{1,t}^{(2)}
:=
\frac{1}{m_t}
\sum_{s\in S_t}
\frac{\pi_{\mathrm{ref}}(A_s\mid X_s)}{\pi_s(A_s\mid X_s)}
\,f_2(\widetilde\mu_{s-1})(O_s),
\]
\[
\widehat\Phi_{2,t}^{(4)}
:=
\frac{1}{m_t}
\sum_{s\in S_t}
\frac{\pi_{\mathrm{ref}}(A_s\mid X_s)}{\pi_s(A_s\mid X_s)}
\,f_4(\widetilde\mu_{s-1})(O_s).
\]
Define the corresponding Cesàro targets
\[
\overline\Phi_{1,t}^{(k)}
:=
\frac{1}{m_t}
\sum_{s\in S_t}
P_{\mathrm{ref}}
\big[
f_k(g_t,\widetilde\mu_{s-1})(O)
\big],
\qquad
k\in\{1,3,5\},
\]
\[
\overline\Phi_{1,t}^{(2)}
:=
\frac{1}{m_t}
\sum_{s\in S_t}
P_{\mathrm{ref}}
\big[
f_2(\widetilde\mu_{s-1})(O)
\big],
\qquad
\overline\Phi_{2,t}^{(4)}
:=
\frac{1}{m_t}
\sum_{s\in S_t}
P_{\mathrm{ref}}
\big[
f_4(\widetilde\mu_{s-1})(O)
\big].
\]
Finally, define the current-time targets
\[
\Phi_{0,1,t}^{(1)}
:=
P_{\mathrm{ref}}
\big[
f_1(g_t,\widetilde\mu_{t-1})(O)
\big],
\qquad
\Phi_{0,1,t}^{(2)}
:=
P_{\mathrm{ref}}
\big[
f_2(\widetilde\mu_{t-1})(O)
\big],
\qquad
\Phi_{0,1,t}^{(3)}
:=
P_{\mathrm{ref}}
\big[
f_3(g_t,\widetilde\mu_{t-1})(O)
\big],
\]
\[
\Phi_{0,2,t}^{(4)}
:=
P_{\mathrm{ref}}
\big[
f_4(\widetilde\mu_{t-1})(O)
\big],
\qquad
\Phi_{0,2,t}^{(5)}
:=
P_{\mathrm{ref}}
\big[
f_5(g_t,\widetilde\mu_{t-1})(O)
\big].
\]

By Lemma~\ref{lem:projected_f_decomposition},
\begin{equation}
\label{eq:M1_piece_decomposition}
\widehat M_{1,t}(w_n)
=
\widehat\Phi_{2,t}^{(4)}
+
\widehat\Phi_{1,t}^{(5)},
\qquad
M_{1,t}(w_n)
=
\Phi_{0,2,t}^{(4)}
+
\Phi_{0,2,t}^{(5)},
\end{equation}
and
\begin{equation}
\label{eq:M2_piece_decomposition}
\widehat M_{2,t}(w_n)
=
\widehat\Phi_{1,t}^{(1)}
+
\widehat\Phi_{1,t}^{(2)}
+
\widehat\Phi_{1,t}^{(3)},
\qquad
M_{2,t}(w_n)
=
\Phi_{0,1,t}^{(1)}
+
\Phi_{0,1,t}^{(2)}
+
\Phi_{0,1,t}^{(3)}.
\end{equation}

\subsection{Proof of Theorem~\ref{thm:projected_variance_consistency}}

\paragraph{Reference law, truncated policy class, and intermediate Ces\`aro moments.}
For any measurable \(h:\cX\times\cA\times\cY\to\mathbb R\), define
\[
P_{\mathrm{ref}}h
:=
\E_{X\sim P_X,\ A\sim \pi_{\mathrm{ref}}(\cdot\mid X),\ Y\sim P_{Y\mid X,A}(\cdot\mid X,A)}
\bigl[h(X,A,Y)\bigr].
\]
Equivalently,
\[
P_{\mathrm{ref}}
=
P_X\otimes \pi_{\mathrm{ref}}(\cdot\mid X)\otimes P_{Y\mid X,A}(\cdot\mid X,A).
\]

For \(\delta>0\), define the truncated policy subclass
\[
\Pi_e(\delta)
:=
\Bigl\{\pi\in\Pi:\inf_{b\in\cA,\ x\in\cX}\pi(b\mid x)\ge \delta\Bigr\},
\]
and its ratio representation
\[
\mathcal G_e(\delta)
:=
\Bigl\{\pi/\pi_{\mathrm{ref}}:\pi\in\Pi_e(\delta)\Bigr\}
\subseteq \mathcal G_{\mathrm{ref}}.
\]

For \(t\ge n+2\), define the intermediate Ces\`aro moments
\[
\overline M_{1,t}(w_n)
:=
\overline\Phi_{2,t}^{(4)}+\overline\Phi_{1,t}^{(5)},
\qquad
\overline M_{2,t}(w_n)
:=
\overline\Phi_{1,t}^{(1)}+\overline\Phi_{1,t}^{(2)}+\overline\Phi_{1,t}^{(3)}.
\]
Then, for \(j\in\{1,2\}\),
\[
\widehat M_{j,t}(w_n)-M_{j,t}(w_n)
=
\bigl(\widehat M_{j,t}(w_n)-\overline M_{j,t}(w_n)\bigr)
+
\bigl(\overline M_{j,t}(w_n)-M_{j,t}(w_n)\bigr).
\]

For \(k\in\{1,3,5\}\) and \(\delta>0\), define the projected localized classes
\[
\mathcal F_{k,t}(\delta)
:=
\Bigl\{
\bigl(f_k(g,\widetilde\mu_{s-1})(O_s)\bigr)_{s\in S_t}
:\,
g\in \mathcal G_e(\delta)
\Bigr\}.
\]

\begin{proof}[Proof of Theorem~\ref{thm:projected_variance_consistency}]
Condition on \(\cF_n\). Then \(w_n\) is deterministic and \(\|w_n\|_{\cH_{\cY}}=1\). Define
\[
\widetilde Y_s:=\langle w_n,\phi_\cY(Y_s)\rangle_{\cH_{\cY}},
\qquad
\widetilde\mu_{s-1}(b,x):=\langle w_n,\widehat\mu^{(2)}_{s-1}(b,x)\rangle_{\cH_{\cY}}.
\]
By contraction,
\[
|\widetilde Y_s|\le \sqrt{\kappa},
\qquad
\sup_{s,b,x}|\widetilde\mu_{s-1}(b,x)|\le B_\mu,
\qquad
\|\widetilde\mu_{t-1}-\widetilde\mu_\infty\|_{L_2(P_X\times\mu_{\cA})}
=
O(t^{-\beta})
\quad\text{a.s.}
\]
Thus the inferential-fold problem is scalar.

Let \(\delta_t:=c_\pi t^{-\alpha}\). For the empirical-process part, consider the classes
\[
\mathcal F_{k,t}(\delta_t)
=
\Bigl\{
\bigl(f_k(g,\widetilde\mu_{s-1})(O_s)\bigr)_{s\in S_t}
:\,
g\in\mathcal G_e(\delta_t)
\Bigr\},
\qquad k\in\{1,3,5\}.
\]
The same monotone bracketing argument as in \citep{bibaut2021post}\ applies after projection: the map
\(u\mapsto \langle w_n,u\rangle\) is a contraction, so projection does not enlarge the entropy of the
policy-ratio class. Hence the martingale empirical-process terms satisfy
\[
\bigl|\widehat M_{j,t}(w_n)-\overline M_{j,t}(w_n)\bigr|
=
O_{\mathrm{a.s.}}(m_t^{-\nu_{\mathrm{emp}}})
\]
for some \(\nu_{\mathrm{emp}}>0\), \(j=1,2\).

For the approximation part, use the projected analogue of Lemma~5 of \citet{bibaut2021post}:
for any \(g,g_1\in\mathcal G_e(\delta)\) and any scalar nuisances \(m,m_1\),
\[
|\Phi_{0,1}^{(1)}(g,m)-\Phi_{0,1}^{(1)}(g_1,m_1)|
\lesssim
\delta^{-2}\|g-g_1\|_{L_1(P_X\times\mu_{\cA})}
+
\delta^{-1}\|m-m_1\|_{L_1(P_X\times\mu_{\cA})},
\]
while the remaining four pieces are Lipschitz in \(m\) with no worse than \(O(1)\) constant.
Since \(\overline\Phi\) and \(\Phi_0\) are evaluated at the same current policy \(g_t\), only the nuisance
difference remains. Therefore
\[
\bigl|\overline M_{j,t}(w_n)-M_{j,t}(w_n)\bigr|
\lesssim
\frac{1}{m_t}
\sum_{s\in S_t}
\delta_t^{-1}
\|\widetilde\mu_{s-1}-\widetilde\mu_{t-1}\|_{L_1(P_X\times\mu_{\cA})}
=
O_{\mathrm{a.s.}}(t^{-(\beta-\alpha)}),
\]
because \(\alpha<\beta\).

Let
\[
\nu:=\min\{\nu_{\mathrm{emp}},\,\beta-\alpha\}>0.
\]
Then
\[
\bigl|\widehat M_{j,t}(w_n)-M_{j,t}(w_n)\bigr|
=
O_{\mathrm{a.s.}}(m_t^{-\nu}),
\qquad j=1,2.
\]
Averaging over \(t\in\mathcal I_2\), and using \(m_t=t-n-1\),
\[
\frac1n\sum_{t\in\mathcal I_2} m_t^{-\nu}
=
\frac1n\sum_{u=1}^{n-1}u^{-\nu}
=
O(n^{-\nu'})
\]
for some \(\nu'>0\). Renaming \(\nu'\) as \(\nu(\alpha,\beta,p)\) yields
\eqref{eq:average_M1_consistency} and \eqref{eq:average_M2_consistency}.

Finally,
\[
\widehat\sigma_t^2(w_n)=\bigl(\widehat M_{2,t}(w_n)-\widehat M_{1,t}(w_n)^2\bigr)_+,
\qquad
\sigma_{0,t}^2(w_n)=\bigl(M_{2,t}(w_n)-M_{1,t}(w_n)^2\bigr)_+.
\]
Since \(x\mapsto x_+\) is \(1\)-Lipschitz and \(M_{1,t}(w_n)=\theta_n(a,a')\) is uniformly bounded,
while \(\widehat M_{1,t}(w_n)\to M_{1,t}(w_n)\) a.s., the map \((m_1,m_2)\mapsto (m_2-m_1^2)_+\) is
eventually Lipschitz on the relevant set. Hence \eqref{eq:average_sigma_consistency} follows.
\end{proof}

\section{PROOFS FOR VALIDITY, WITNESS CONSISTENCY, AND POWER}
\label{app:witness_power}

This appendix contains the proofs for the results in
Section~\ref{section:practical_test_power}. We first prove asymptotic validity by combining the null theorem with the variance-consistency result. We then prove consistency of the pilot witness and of the pilot direction. Finally, we prove consistency of the adaptive test under fixed alternatives.

\subsection{Proof of Corollary~\ref{cor:projected_test_validity}}

\begin{proof}[Proof of Corollary~\ref{cor:projected_test_validity}]
This is immediate from Theorem~\ref{thm:projected_test_clt}.
\end{proof}

\subsection{Proof of Theorem~\ref{thm:pilot_witness_consistency}}

\begin{proof}[Proof of Theorem~\ref{thm:pilot_witness_consistency}]
Recall
\[
v_n
=
\frac{1}{n}\sum_{t\in \mathcal I_1}\hat\phi_t^{(1)}.
\]
By the doubly robust identity, for every \(t\in\mathcal I_1\),
\[
\E\!\left[
\hat\phi_t^{(1)}
\,\middle|\,
\cF_{t-1}
\right]
=
\Psi(a,a').
\]
Therefore
\begin{equation}
\label{eq:pilot_martingale_decomp_appendix}
v_n-\Psi(a,a')
=
\frac{1}{n}
\sum_{t\in \mathcal I_1}
\Delta_t,
\qquad
\Delta_t
:=
\hat\phi_t^{(1)}-\E[\hat\phi_t^{(1)}\mid\cF_{t-1}],
\end{equation}
and \((\Delta_t)_{t\in \mathcal I_1}\) is a sequence of
\(\cH_{\cY}\)-valued martingale differences.

It remains to bound \(\E[\|\Delta_t\|_{\cH_{\cY}}^2]\). Since
\[
\|\Delta_t\|_{\cH_{\cY}}^2
\le
2\|\hat\phi_t^{(1)}\|_{\cH_{\cY}}^2
+
2\|\Psi(a,a')\|_{\cH_{\cY}}^2,
\]
it is enough to bound \(\E[\|\hat\phi_t^{(1)}\|_{\cH_{\cY}}^2]\). By the same argument as in Lemma~\ref{lem:projected_qth_moment_bound} with \(q=2\) and \(r=1\) in place of \(r=2\),
\[
\E\!\left[
\|\hat\phi_t^{(1)}\|_{\cH_{\cY}}^2
\,\middle|\,
\cF_{t-1}
\right]
\lesssim t^\alpha.
\]
Hence
\[
\E\!\left[
\|\Delta_t\|_{\cH_{\cY}}^2
\right]
\lesssim t^\alpha.
\]
Martingale orthogonality then yields
\begin{align}
\E\!\left[
\|v_n-\Psi(a,a')\|_{\cH_{\cY}}^2
\right]
&=
\frac{1}{n^2}
\sum_{t\in \mathcal I_1}
\E\!\left[
\|\Delta_t\|_{\cH_{\cY}}^2
\right]
\nonumber\\
&\lesssim
\frac{1}{n^2}
\sum_{t=1}^n t^\alpha
\asymp n^{\alpha-1}.
\label{eq:pilot_rate_appendix}
\end{align}
This proves \eqref{eq:pilot_l2_consistency}. Since \(\alpha<1\), the right-hand side tends to zero, so
\[
v_n\xrightarrow[]{L_2(\cH_{\cY})}\Psi(a,a')
\qquad\text{and hence}\qquad
v_n\xrightarrow[]{p}\Psi(a,a').
\qedhere
\]
\end{proof}

\subsection{Proof of Corollary~\ref{cor:pilot_direction_consistency}}

\begin{proof}[Proof of Corollary~\ref{cor:pilot_direction_consistency}]
Since \(\Psi(a,a')\neq0\), Theorem~\ref{thm:pilot_witness_consistency} implies
\[
v_n\xrightarrow[]{p}\Psi(a,a').
\]
In particular,
\[
\Pr(v_n=0)
\le
\Pr\!\left(
\|v_n-\Psi(a,a')\|_{\cH_{\cY}}
\ge
\|\Psi(a,a')\|_{\cH_{\cY}}
\right)
\longrightarrow 0.
\]
On the event \(\{v_n\neq0\}\), we have \(w_n=v_n/\|v_n\|_{\cH_{\cY}}\). Since the map
\[
v\mapsto \frac{v}{\|v\|_{\cH_{\cY}}}
\]
is continuous away from \(0\), the continuous mapping theorem gives
\[
\frac{v_n}{\|v_n\|_{\cH_{\cY}}}
\xrightarrow[]{p}
\frac{\Psi(a,a')}{\|\Psi(a,a')\|_{\cH_{\cY}}}.
\]
Because \(\Pr(v_n=0)\to0\), the tie-breaker case is asymptotically negligible, and therefore
\[
w_n
\xrightarrow[]{p}
\frac{\Psi(a,a')}{\|\Psi(a,a')\|_{\cH_{\cY}}}.
\]
\end{proof}

\subsection{Proof of Corollary~\ref{cor:projected_test_power}}

\begin{proof}[Proof of Corollary~\ref{cor:projected_test_power}]
Let
\[
\theta_n(a,a')
:=
\langle w_n,\Psi(a,a')\rangle_{\cH_{\cY}}.
\]
By Corollary~\ref{cor:pilot_direction_consistency},
\begin{equation}
\label{eq:theta_positive_limit_appendix}
\theta_n(a,a')
\xrightarrow[]{p}
\|\Psi(a,a')\|_{\cH_{\cY}}
>0.
\end{equation}

Write
\begin{align}
T_n(w_n)
&=
\frac{1}{\sqrt n}
\sum_{t\in\mathcal I_2}
\widetilde\sigma_t(w_n)^{-1}
\Big(
\xi_t(w_n)-\theta_n(a,a')
\Big)
\nonumber\\
&\quad+
\frac{\theta_n(a,a')}{\sqrt n}
\sum_{t\in\mathcal I_2}
\widetilde\sigma_t(w_n)^{-1}.
\label{eq:power_decomp}
\end{align}

We first control the stochastic term. Since
\[
\E\!\left[
\xi_t(w_n)-\theta_n(a,a')
\,\middle|\,
\cF_{t-1}
\right]
=0,
\]
the first term in \eqref{eq:power_decomp} is a martingale array with predictable variation
\[
\frac{1}{n}
\sum_{t\in\mathcal I_2}
\frac{\sigma_{0,t}^2(w_n)}{\widetilde\sigma_t^2(w_n)}
\xrightarrow[]{p}1
\]
by Corollary~\ref{assumption:average_inverse_sd_consistency}. Thus the first term in \eqref{eq:power_decomp} is \(O_p(1)\).

We now lower bound the deterministic drift. For \(s<t\) in \(\mathcal I_2\), boundedness of the kernel, the nuisance envelope, and Assumption~\ref{assumption:exploration_floor} give
\[
|\check\xi_{s,t}(w_n)|
\lesssim
\frac{\mathbbm 1\{A_s=a\}}{\pi_t(a\mid X_s)}
+
\frac{\mathbbm 1\{A_s=a'\}}{\pi_t(a'\mid X_s)}
+
1
\lesssim t^\alpha.
\]
Moreover,
\[
w_{s,t}
=
\frac{\pi_t(A_s\mid X_s)}{\pi_s(A_s\mid X_s)}
\lesssim s^\alpha
\lesssim t^\alpha,
\]
since \(\pi_t(A_s\mid X_s)\le 1\) and \(\pi_s(A_s\mid X_s)\gtrsim s^{-\alpha}\). Therefore
\[
\widehat M_{2,t}(w_n)
\lesssim t^\alpha \cdot t^{2\alpha}
=
t^{3\alpha},
\]
so
\[
\widehat\sigma_t(w_n)\lesssim t^{3\alpha/2}
\qquad\text{and hence}\qquad
\widetilde\sigma_t(w_n)^{-1}\gtrsim t^{-3\alpha/2}.
\]
Consequently,
\[
\frac{1}{\sqrt n}
\sum_{t\in\mathcal I_2}
\widetilde\sigma_t(w_n)^{-1}
\gtrsim
\frac{1}{\sqrt n}
\sum_{t=n+1}^{2n}
t^{-3\alpha/2}
\asymp
n^{1/2-3\alpha/2}.
\]
By Assumption~\ref{assumption:exploration_floor_strong}, \(\alpha<\alpha_\star(\beta,p)\le 1/3\), so \(1/2-3\alpha/2>0\). Therefore the drift term in \eqref{eq:power_decomp} diverges to \(+\infty\) in probability by \eqref{eq:theta_positive_limit_appendix}, while the stochastic term remains \(O_p(1)\). Hence
\[
T_n(w_n)\xrightarrow[]{p}+\infty.
\]
This proves consistency of the rejection rule.
\end{proof}

\section{CLOSED FORMS FOR SAMPLE-SPLIT AND PROJECTED ADAPTIVE STATISTICS}
\label{appendix:ClosedFormDerivation}

\subsection{Projected adaptive DR-KTE (PADR-KTE)}
\label{appendix:padr_kte_closed_form}

We now give the kernel-matrix closed form for the projected adaptive statistic
introduced in Sections~\ref{section:adaptive_test}--\ref{section:practical_test_power}.
Unlike the previous cross-fold variance-stabilized construction, the new estimator is
not bilinear in two stabilized folds. The first fold is used only to construct the RKHS
witness, while the second fold is reduced to a scalar sequential adaptive-DR problem
after projection.

For simplicity we write the binary contrast \(a=1\) versus \(a'=0\). For a general
pair \((a,a')\), replace the indicators and propensities accordingly.

\paragraph{Chronological split and fold-local indexing.}
Write \(T=2n\) and split chronologically:
\[
\cI_1=\{1,\dots,n\},\qquad
\cI_2=\{n+1,\dots,2n\}.
\]
For convenience, index each fold locally by \(u=1,\dots,n\):
\[
(X_{1,u},A_{1,u},Y_{1,u}) := (X_u,A_u,Y_u),
\qquad
(X_{2,u},A_{2,u},Y_{2,u}) := (X_{n+u},A_{n+u},Y_{n+u}).
\]
Let \(\pi_u^{(1)}(\cdot\mid x):=\pi_u(\cdot\mid x)\) and
\(\pi_u^{(2)}(\cdot\mid x):=\pi_{n+u}(\cdot\mid x)\).

Let \(k_\cY\) be a PD kernel on outcomes with RKHS \(\cH_\cY\) and feature map
\(\varphi_\cY(y)=k_\cY(\cdot,y)\).
For a fold \(r\in\{1,2\}\), define the feature operator
\[
\Phi_{\cY,r} c
=
\sum_{u=1}^n c_u\,\varphi_\cY(Y_{r,u}) \in \cH_\cY,
\qquad
\langle \Phi_{\cY,r} c,\Phi_{\cY,r'} d\rangle_{\cH_\cY}
=
c^\top K_{\cY}^{(r,r')} d,
\]
where \(K_{\cY}^{(r,r')}=[k_\cY(Y_{r,u},Y_{r',v})]_{u,v=1}^n\).
Let \(K_{\cX}^{(r,r)}=[k_\cX(X_{r,u},X_{r,v})]_{u,v=1}^n\) be the within-fold covariate Gram block.

\paragraph{Past-only KRR coefficient vectors.}
Fix a fold \(r\in\{1,2\}\), a local time \(u\in\{1,\dots,n\}\), and an arm \(b\in\{0,1\}\).
Define the past arm-specific index set
\[
J_{r,u}^{(b)}
:=
\{\,v\in\{1,\dots,u-1\}:A_{r,v}=b\,\}.
\]
Let \(e_u\in\mathbb R^n\) be the \(u\)-th canonical basis vector, and let
\(S_{r,u}^{(b)}\in\mathbb R^{n\times |J_{r,u}^{(b)}|}\) be the selector that inserts a vector
indexed by \(J_{r,u}^{(b)}\) into the full fold coordinates.
With ridge \(\lambda>0\), define the past-only KRR coefficient vector
\[
h_{r,u}^{(b)}
:=
S_{r,u}^{(b)}
\Big(
K_{\cX}^{(r,r)}[J_{r,u}^{(b)},J_{r,u}^{(b)}]+\lambda I
\Big)^{-1}
K_{\cX}^{(r,r)}[J_{r,u}^{(b)},u]
\in\mathbb R^n,
\]
with the convention \(h_{r,u}^{(b)}=0\) if \(J_{r,u}^{(b)}=\emptyset\).
Then
\[
\widehat\mu_{u-1}^{(r)}(b,X_{r,u})
=
\Phi_{\cY,r}\,h_{r,u}^{(b)}.
\]

\paragraph{Fold-1 pilot witness.}
For the contrast \(1-0\), the fold-1 RKHS score at local time \(u\) is
\[
\hat\phi_u^{(1)}
=
\Phi_{\cY,1}\,d_u^{\mathrm{pil}},
\]
where
\[
d_u^{\mathrm{pil}}
=
\frac{\mathbf 1\{A_{1,u}=1\}}{\pi_u^{(1)}(1\mid X_{1,u})}
\bigl(e_u-h_{1,u}^{(1)}\bigr)
-
\frac{\mathbf 1\{A_{1,u}=0\}}{\pi_u^{(1)}(0\mid X_{1,u})}
\bigl(e_u-h_{1,u}^{(0)}\bigr)
+
h_{1,u}^{(1)}-h_{1,u}^{(0)}.
\]
Hence
\[
\bar d_1
:=
\frac1n\sum_{u=1}^n d_u^{\mathrm{pil}},
\qquad
v_n
=
\Phi_{\cY,1}\bar d_1,
\qquad
\|v_n\|_{\cH_\cY}^2
=
\bar d_1^\top K_{\cY}^{(1,1)}\bar d_1.
\]
If \(v_n\neq 0\), define the witness coefficient vector
\[
c^{\mathrm{wit}}
:=
\frac{\bar d_1}{\sqrt{\bar d_1^\top K_{\cY}^{(1,1)}\bar d_1}},
\qquad
w_n=\Phi_{\cY,1}c^{\mathrm{wit}}.
\]
Then the projected scalar outcomes on fold 2 are
\[
z
:=
K_{\cY}^{(2,1)} c^{\mathrm{wit}}
\in\mathbb R^n,
\qquad
z_u
=
\langle w_n,\varphi_\cY(Y_{2,u})\rangle_{\cH_\cY}.
\]
If \(v_n=0\), replace \(w_n\) by the fixed tie-breaker \(u_0\) and compute
\(z_u=\langle u_0,\varphi_\cY(Y_{2,u})\rangle_{\cH_\cY}\).

\paragraph{Fold-2 projected scalar nuisances.}
After projection, the inferential fold is scalar. Using the same past-only KRR vectors
\(h_{2,u}^{(0)},h_{2,u}^{(1)}\), define the scalar nuisance evaluations
\[
m_u^{(b)}
:=
z^\top h_{2,u}^{(b)}
=
\langle w_n,\widehat\mu_{u-1}^{(2)}(b,X_{2,u})\rangle_{\cH_\cY},
\qquad b\in\{0,1\}.
\]

\paragraph{Current projected score.}
The projected DR score at inferential local time \(u\in\{1,\dots,n\}\) is
\[
\xi_u
=
z^\top c_u,
\]
where
\[
c_u
=
\frac{\mathbf 1\{A_{2,u}=1\}}{\pi_u^{(2)}(1\mid X_{2,u})}
\bigl(e_u-h_{2,u}^{(1)}\bigr)
-
\frac{\mathbf 1\{A_{2,u}=0\}}{\pi_u^{(2)}(0\mid X_{2,u})}
\bigl(e_u-h_{2,u}^{(0)}\bigr)
+
h_{2,u}^{(1)}-h_{2,u}^{(0)}.
\]
Equivalently,
\[
\xi_u
=
\frac{\mathbf 1\{A_{2,u}=1\}}{\pi_u^{(2)}(1\mid X_{2,u})}
\bigl(z_u-m_u^{(1)}\bigr)
-
\frac{\mathbf 1\{A_{2,u}=0\}}{\pi_u^{(2)}(0\mid X_{2,u})}
\bigl(z_u-m_u^{(0)}\bigr)
+
m_u^{(1)}-m_u^{(0)}.
\]

\paragraph{History-recycled projected scores.}
For \(1\le s<t\le n\), the historical nuisance at time \(s\) is kept fixed, while only the
evaluation-time propensity is changed to \(\pi_t^{(2)}\). Define
\[
\check\xi_{s,t}
=
z^\top c_{s,t},
\]
where
\[
c_{s,t}
=
\frac{\mathbf 1\{A_{2,s}=1\}}{\pi_t^{(2)}(1\mid X_{2,s})}
\bigl(e_s-h_{2,s}^{(1)}\bigr)
-
\frac{\mathbf 1\{A_{2,s}=0\}}{\pi_t^{(2)}(0\mid X_{2,s})}
\bigl(e_s-h_{2,s}^{(0)}\bigr)
+
h_{2,s}^{(1)}-h_{2,s}^{(0)}.
\]
Equivalently,
\[
\check\xi_{s,t}
=
\frac{\mathbf 1\{A_{2,s}=1\}}{\pi_t^{(2)}(1\mid X_{2,s})}
\bigl(z_s-m_s^{(1)}\bigr)
-
\frac{\mathbf 1\{A_{2,s}=0\}}{\pi_t^{(2)}(0\mid X_{2,s})}
\bigl(z_s-m_s^{(0)}\bigr)
+
m_s^{(1)}-m_s^{(0)}.
\]

\paragraph{Sequential conditional variance estimator.}
For \(1\le s<t\le n\), define the importance ratio
\[
\rho_{s,t}
:=
\frac{\pi_t^{(2)}(A_{2,s}\mid X_{2,s})}
{\pi_s^{(2)}(A_{2,s}\mid X_{2,s})}.
\]
For a fixed inferential time \(t\ge 2\), collect the recycled score coefficients into
\[
C_t
:=
\bigl[c_{1,t},\dots,c_{t-1,t}\bigr]
\in\mathbb R^{n\times (t-1)},
\qquad
\rho_t
:=
(\rho_{1,t},\dots,\rho_{t-1,t})^\top.
\]
Then
\[
\check\xi_t
:=
(\check\xi_{1,t},\dots,\check\xi_{t-1,t})^\top
=
C_t^\top z,
\]
and the sequential moments are
\[
\widehat M_{1,t}
=
\frac{1}{t-1}\,\rho_t^\top C_t^\top z,
\qquad
\widehat M_{2,t}
=
\frac{1}{t-1}\,\rho_t^\top \bigl(C_t^\top z\bigr)^{\odot 2},
\qquad t\ge 2.
\]
Set the initial convention
\[
\widehat\sigma_1^2:=1.
\]
For \(t\ge 2\), define
\[
\widehat\sigma_t^2
=
\Big(
\widehat M_{2,t}-\widehat M_{1,t}^2
\Big)_+,
\qquad
\widetilde\sigma_t
=
\sqrt{\widehat\sigma_t^2}\vee \epsilon_n.
\]
For \(t=1\), set \(\widetilde\sigma_1:=1\vee \epsilon_n\).

\paragraph{PADR-KTE statistic.}
The projected adaptive DR-KTE statistic is
\[
T_{\mathrm{PADR\text{-}KTE}}
=
\frac{1}{\sqrt n}
\sum_{t=1}^n
\widetilde\sigma_t^{-1}\,\xi_t.
\]

\paragraph{Efficient evaluation.}
The vectors \(h_{1,u}^{(b)}\) and \(h_{2,u}^{(b)}\) are computed once from covariate Gram blocks.
The pilot fold enters the inferential fold only through the witness coefficients \(c^{\mathrm{wit}}\)
and the projected outcome vector \(z=K_{\cY}^{(2,1)}c^{\mathrm{wit}}\).
After that point, all inferential computations are scalar:
only the evaluation-time denominators \(\pi_t^{(2)}(b\mid X_{2,s})\) and the ratios \(\rho_{s,t}\)
vary with \(t\); all historical smoother vectors \(h_{2,s}^{(b)}\) are reused.

\section{SUPPLEMENTARY ON NUMERICAL SIMULATIONS}
\label{app:additional_experiments}
This section provides a detailed supplement to the numerical simulations presented in Section \ref{section:experiments}. We first specify the kernel function leveraged in our method. Following this, we discuss the baseline algorithms against which our approach was compared, and conclude by detailing additional experimental setups and presenting supplementary numerical results.

\subsection{Kernel}
In our experiments, we employed the Gaussian kernel (also known as the Radial Basis Function or RBF kernel), defined for all $h_i, h_j \in \mathbb{R}^{d_\cH}$ as:
\begin{align*}
    k_\cH(h_i, h_j) = \exp\left(-\frac{\|h_i - h_j\|_2^2}{2 \gamma^2}\right).
\end{align*}
The parameter $\gamma$ is the length-scale of the kernel, which controls the smoothness of the resulting function space. The Gaussian kernel is widely used in practice and satisfies the crucial properties of boundedness, continuity, and characteristicity \citep{SriGreFukLanetal10}. For both the covariate space $\cX$ and the outcome space $\cY$, we utilized the Gaussian kernel, setting the length-scales based on the median of the pairwise Euclidean distances from the given data. Specifically, for a dataset $\{h_i\}_{i=1}^T$, the median pairwise distance is given by
\begin{align*}
    \gamma_{\text{median}} = \text{median}\{\|h_i - h_j\|_2 \mid 1 \le i < j \le T\}.
\end{align*}
In particular, we chose the length-scale for the covariate kernel ($k_\cX$) to be equal to the median pairwise distance, and for the outcome kernel ($k_\cY$), we set the length-scale to be one half of the calculated median distance.
\subsection{Baselines}

\paragraph{(i) CADR (Contextual Adaptive Doubly Robust):}
CADR is a \emph{stabilized DR} estimator specifically designed for data that is both contextual (dependent on covariates $X$) and adaptively collected (where the data collection process changes over time). The estimator operates by forming a canonical gradient \(D'(g_t,\bar Q_{t-1})(X_t,A_t,Y_t)\)—a term that incorporates both the policy and an outcome model. This gradient is then aggregated across time using history-measurable inverse standard-deviation weights, \(\hat\sigma_t^{-1}\). The components are defined as follows:
\begin{itemize}
\item $g_t$ is the logging policy at time $t$.
    \item $\bar{Q}_{t}: \cA \times \cX \rightarrow \cY$: An estimate of the Conditional Outcome Model $\mathbb{E}[Y \mid A = \cdot, X = \cdot]$. Crucially, for every $t$, $\bar{Q}_{t}$ is trained using only data observed up to time $t$.
    \item \(\hat\sigma_t^{-1}\): The inverse of \(\hat\sigma_t\), which estimates the conditional standard deviation $\sigma_{0, t} = \text{Var}\!\big(D'(g_t,\bar{Q}_{t-1})(O_t)\mid O_{1:t-1}\big)^{1/2}$. These weights stabilize the variance of the overall estimate.
    \item $O_t$: The set of observed variables at time $t$, $O_t = (X_t, A_t, Y_t)$, and $O_{1:t-1} = (O(1), \ldots, O(t-1))$.
\end{itemize}

The stabilized estimate is constructed as
\[
\widehat\Psi_T
=\Big(\tfrac{1}{T}\sum_{t=1}^T \hat\sigma_t^{-1}\Big)^{-1}
\cdot \tfrac{1}{T}\sum_{t=1}^T \hat\sigma_t^{-1}\,D'\!\big(g_t,\bar Q_{t-1}\big)(O_t),
\]
with asymptotic normality under consistency of the \emph{conditional} standard-deviation
estimators \(\hat\sigma_t\) (each trained on past data only) and a mild exploration condition
(\(g_t(a\mid x)\gtrsim t^{-1/2}\)); see \citep[Algorithm~1; Theorem~1; Section~3]{bibaut2021post}.%
\footnote{CADR constructs \(\hat\sigma_t^2\) via importance-reweighting across past policies
\(g_s\) using ratios \(g_t/g_s\) and proves almost-sure consistency of \(\hat\sigma_t^2\) under
a bracketing-entropy bound on the logging policy class and a rate for the outcome-regression
sequence \(\bar Q_t\).} We implement CADR exactly as specified with fold-wise, predictable
nuisance fits and \(\hat\sigma_t\) built from past data only.

\textbf{(ii) Variance-stabilized AIPW of \citet{hadad2021confidence}.}
\citet{hadad2021confidence} propose an \emph{adaptively-weighted AIPW} family for non-contextual
adaptive experiments that ensures martingale variance convergence via \emph{variance-stabilizing
weights}. Let \(\Gamma_t\) denote the (A)IPW score for a fixed arm and \(e_t\) its propensity.
Weights \(\{h_t\}\) are chosen so that \(\sum_{t} h_t^2/e_t\) is \emph{deterministic} (stick-breaking),
which yields a studentized statistic with a standard normal limit. Two named allocation schemes are:
\emph{constant allocation} \(\lambda_t^{\mathrm{const}}=\frac{1}{T-t+1}\) (giving \(h_t\propto \sqrt{e_t/T}\)),
and the \emph{two-point allocation} \(\lambda_t^{\mathrm{two\mbox{-}point}}\) that interpolates between
high-propensity and vanishing-propensity regimes using a heuristic for future propensities; both
satisfy the sufficient bounds of their Theorem~3. We implement this baseline as
\textbf{AW-AIPW (Hadad)} with both \texttt{constant} and \texttt{two\_point} allocation options,
and with AIPW scores; see \citep[Section~2.2--2.3; Theorem~2--3; Equation~(12)--(18)]{hadad2021confidence}. 

\subsection{Additional description of the experiments}

In this Appendix, we provide additional details and descriptions for the experiments in our main text.

\subsubsection{Synthetic data}\label{app:synthetic}

All data (covariates, treatments, responses) is simulated. Each round draws a context $X_t \in \mathbb{R}^5$ i.i.d.\ from $\mathcal{N}(0,I_5)$. We consider three cases for the underlying function $f$ that generates the potential outcome: 
\begin{itemize}
    \item[(i)] \emph{cosine model} with $f(x)=\text{cos}(\beta^\top x)$ and $\beta=(0.1,0.2,0.3,0.4,0.5)$;
    \item[(ii)] \emph{linear model} with $f(x)=\beta^\top x$ and the same $\beta$; and
    \item[(iii)] \emph{sigmoidal model} with $f(x)=\sigma(\beta^\top x)$ where $\sigma(z)=ln(|16z - 8| + 1) \cdot \text{sign}(z - 0.5)$ and the same $\beta$.
\end{itemize}
 Then, potential outcomes are generated as $Y_t(0)=f(X_t)+\varepsilon_t$ and $Y_t(1)=f(X_t)+\delta_t+\varepsilon_t$, with i.i.d.\ noise $\varepsilon_t\sim \mathcal{N}(0,0.5)$. 

\paragraph{Scenarios.}
We use the four scenarios of \citet{martinez2023} through the treatment effect $\delta_t$: Scenario~I (null) uses $\delta_t=0$; Scenario~II (mean shift) uses $\delta_t=2$; Scenario~III (symmetric mixture) uses $\delta_t=2\,S_t$ with $S_t\in\{-1,+1\}$ Rademacher$(0.5)$; Scenario~IV (random scale) uses $\delta_t\sim \mathrm{Uniform}[-4,4]$. These match the no-effect, constant-mean, symmetric mixture, and random-scale shifts respectively with exact constant values in \citep{martinez2023}.

\paragraph{Adaptive data collection (two arms, $\varepsilon$-greedy with online ridge).}
Each arm $a\in\{0,1\}$ maintains an online ridge model for the potential outcome $Y_t(a)$ based on an augmented design vector $x_t^{\mathrm{aug}}=(1,X_t)$ that includes an unpenalized intercept.  
The ridge state for each arm is a pair $(S_a,b_a)$, where $S_a\in\mathbb{R}^{6\times6}$ is initialized as
\[
S_a = \mathrm{diag}(0,\lambda,\ldots,\lambda), \qquad b_a = 0,
\]
with $\lambda=10^{-2}$ applied to the $d=5$ non-bias coordinates.  
At each round $t$, the current model parameters are updated by solving the linear system
\[
S_a \theta_a = b_a, \qquad a\in\{0,1\},
\]
yielding the estimated regression weights $\theta_a$. The predicted rewards are
\[
q_a(t) = \langle \theta_a,\, x_t^{\mathrm{aug}} \rangle, \qquad a\in\{0,1\}.
\]

\smallskip
The exploration probability decays with time according to
\[
\varepsilon_t = \max\!\big(\varepsilon_{\min},\, \varepsilon_0/(t+1)^p\big),
\quad \text{with } \varepsilon_0 = 0.2,\ \varepsilon_{\min} = 0.05,\ p = 0.99.
\]
Given $(q_0(t),q_1(t))$, the $\varepsilon$-greedy decision rule defines the logging propensities as
\[
\pi_t(1\mid X_t)=
\begin{cases}
1 - \tfrac{1}{2}\varepsilon_t, & q_1(t)>q_0(t),\\[4pt]
\tfrac{1}{2}\varepsilon_t, & q_1(t)<q_0(t),\\[4pt]
0.5, & q_1(t)=q_0(t),
\end{cases}
\qquad
\pi_t(0\mid X_t) = 1 - \pi_t(1\mid X_t).
\]
An action $A_t\in\{0,1\}$ is then sampled according to these propensities, and the observed reward is $Y_t = Y_t(A_t)$.  
The scalar weight used in subsequent estimators is the realized propensity,
\[
w_t =
\begin{cases}
\pi_t(1\mid X_t), & A_t=1,\\[4pt]
\pi_t(0\mid X_t), & A_t=0.
\end{cases}
\]

\smallskip
After observing $(X_t,A_t,Y_t)$, only the chosen arm’s ridge state is updated as
\[
S_{A_t} \leftarrow S_{A_t} + x_t^{\mathrm{aug}} (x_t^{\mathrm{aug}})^\top, 
\qquad
b_{A_t} \leftarrow b_{A_t} + x_t^{\mathrm{aug}} Y_t.
\]
This sequential rule generates a non-i.i.d.\ adaptive trajectory with time-varying propensities $\pi_t(1\mid X_t)$ that progressively concentrate as the regression parameters stabilize.

\paragraph{Propensity matrices for foldwise evaluation.}
For test statistics that require foldwise policy-on-fold propensities, we snapshot $\theta_a$ over time to build matrices that map each decision time to propensities evaluated on all contexts within the same fold. Concretely, we split the trajectory into two non-adaptive folds using the default \emph{alternating} split (odd vs.\ even indices, chronological within each). For each fold $r$ and each in-fold time $t$, we compute $\pi_t(1\mid X_s)$ for all in-fold contexts $X_s$ using the $\theta_a$ snapshot at time $t$, yielding dense $|\mathcal{I}_r|\times|\mathcal{I}_r|$ propensity matrices per fold (with the same greedy/non-greedy/tie rule as above). These matrices, together with the realized $w_t$, are passed to the test procedures.

\paragraph{Kernels and run lengths.}
Outcome similarities use an RBF kernel with bandwidth set as $\gamma=1/\sigma^2$ (i.e., $\gamma=2.0$ when $\sigma^2=0.5$), unless otherwise stated. Each experiment uses a trajectory length $T=1000$ and we run $200$ Monte-Carlo replications per configuration. All other defaults follow the description above.

\subsubsection{IHDP data}

To evaluate our proposed method on a real-world benchmark, we generate a semi-synthetic dataset based on the Infant Health and Development Program (IHDP) data \citep{Hill01012011}. The original IHDP data originates from a randomized experiment on the effects of specialist home visits on cognitive test scores.

Following the preprocessing steps used in \citep{martinez2023}, we retain $908$ samples with $18$ covariates ($9$ continuous, $9$ categorical), resulting in $X_t \in \mathbb{R}^{18}$ for all $t$. We synthesize the adaptive policies $\pi_t$ with two arms, using an $\epsilon$-greedy with online ridge regression. This policy structure is identical to the one discussed in the preceding section, and it results in binary treatments, $A_t \in \{0, 1\}$. 

The potential outcomes are generated according to the following equations:
\begin{align*}
    Y_t(0) &= \text{cos}(\beta^\top X_t) + \epsilon_t, \quad Y_t(1) = \text{cos}(\beta^\top X_t) + \delta_t + \epsilon_t.
\end{align*}
Here, the term $\delta_t$ is used to control the treatment effect, defining four different experimental scenarios. The noise term $\epsilon_t \sim \mathcal{N}(0, 0.5)$ is an i.i.d. Gaussian random variable with zero mean and variance $0.5$, i.e., $\epsilon_t \sim \mathcal{N}(0, 0.5)$.

\paragraph{Scenarios.}
We utilize the same four scenarios, that we adapted for the synthetic data experiments from \citet{martinez2023}, by defining the treatment effect term $\delta_t$: (i) Scenario I (Null): The treatment has no effect, defined by $\delta_t=0$; (ii) Scenario II (Mean Shift): The treatment introduces a constant positive shift, defined by $\delta_t=2$; (iii) Scenario III (Symmetric Mixture): The treatment effect is a symmetric mixture, defined by $\delta_t=2\,S_t$ with $S_t\in\{-1,+1\}$ Rademacher$(0.5)$; (iv) Scenario IV (Random Scale): The treatment effect is randomly scaled, defined by $\delta_t\sim \mathrm{Uniform}[-4,4]$.

\paragraph{Evaluation protocol.} We evaluated our method's performance across varying sample sizes. This was done by running experiments on the IHDP dataset using subsampling without replacement, where the subset size was varied uniformly within the set 
$\{100, 150, 200, \ldots, 850, 900, 908\}$, with $908$ representing the full available dataset. We utilized the non-adaptive alternating fold splitting protocol, consistent with our synthetic dataset experiments, and ran each distinct experiment over $200$ Monte-Carlo replications. 

For the Gaussian kernels used, we followed a median heuristic: the length-scale for the covariate kernel was set equal to the median pairwise distance, while the length-scale for the outcome kernel was set to one half of that median distance. The regularization parameter $\lambda$ was set to $10^{-2}$.

The true positive rates for Scenarios II-IV, utilizing the full available dataset, are presented in Table \ref{table:IHDP_results_main}. A separate discussion detailing additional results that incorporate varying data sizes is provided in Section \ref{appendix:AdditionalNumericalResults}.

\subsubsection{dSprite dataset}
\label{app:dsprite-details}

\begin{figure}
    \centering
    \includegraphics[width=0.5\linewidth]{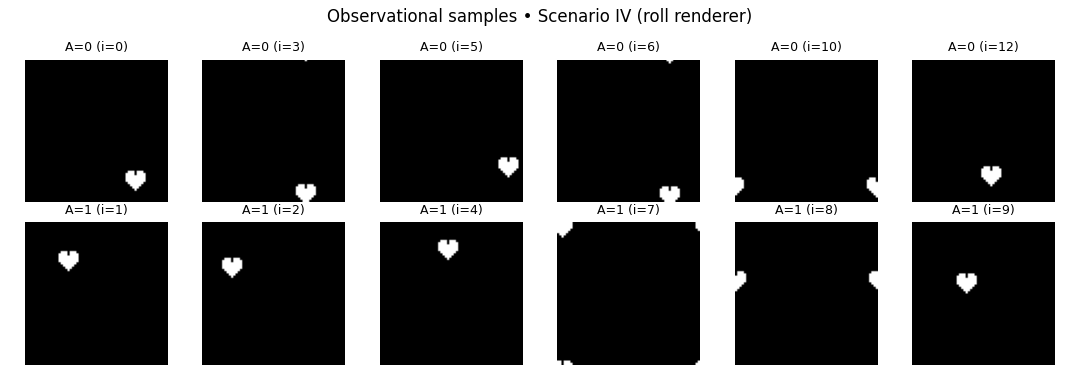}
    \caption{Observational samples from the dSprite data in Scenario IV}
    \label{fig:observatioanl_dsprite}
\end{figure}

\begin{figure}
    \centering
    \includegraphics[width=0.5\linewidth]{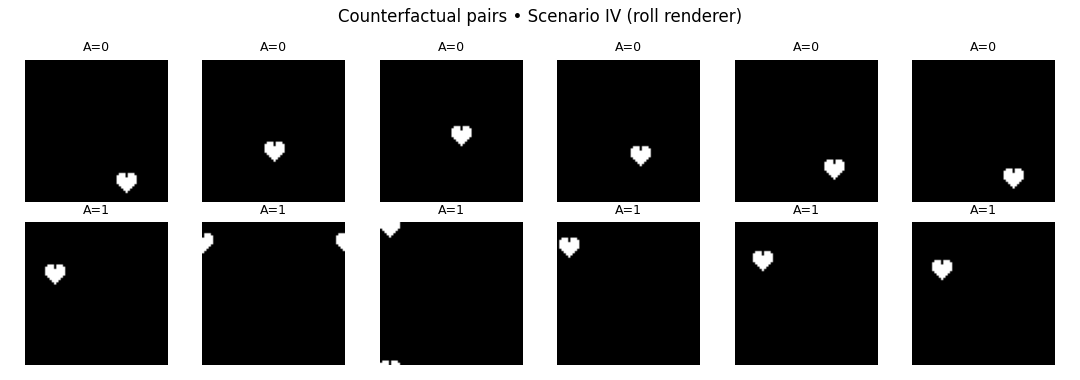}
    \caption{Counterfactual pairs from the dSprite data in Scenario IV}
    \label{fig:counterfactual_dsprite}
\end{figure}

We adapt the structured image benchmark of \citet{xu2023causal} and adapt it to the two-scenario setting of our adaptive kernel test. Each outcome $Y\in[0,1]^{64\times64}$ is a grayscale image of a heart shape on a black background, rendered from latent coordinates $(\mathrm{posX},\mathrm{posY})\in[0,1]^2$. Contexts $X_t=(x_t^{(1)},x_t^{(2)})$ are sampled uniformly from $\mathrm{Unif}([0,1]^2)$, and images are generated through a deterministic renderer
\[
Y_t(a)=g(X_t,a)\in[0,1]^{64\times64},
\]
where $a\in\{0,1\}$ indexes the treatment and $g$ draws a white heart centered at position $(x_t^{(1)}+\Delta_a^{(1)},\,x_t^{(2)}+\Delta_a^{(2)})$ with fixed scale and rotation. The offsets $(\Delta_a^{(1)},\Delta_a^{(2)})$ define the two experimental regimes:
\[
\begin{aligned}
\text{Scenario I (null):}\quad & (\Delta_0^{(1)},\Delta_0^{(2)})=(0,0),\quad (\Delta_1^{(1)},\Delta_1^{(2)})=(0,0);\\[2pt]
\text{Scenario IV (shift):}\quad & (\Delta_0^{(1)},\Delta_0^{(2)})=(0,0),\quad (\Delta_1^{(1)},\Delta_1^{(2)})=(\delta,0),
\end{aligned}
\]
where $\delta=0.15$ induces a rightward translation of the heart under $A=1$ while preserving mean pixel intensity. Gaussian pixel noise $\mathcal{N}(0,0.01)$ is added to each image. Hence, the marginal intensity distributions of $Y(0)$ and $Y(1)$ coincide, but their spatial structure differs. Figure~\ref{fig:observatioanl_dsprite} shows observational samples generated under Scenario~IV, where the adaptive policy produces trajectories with spatially translated outcomes. 
Figure~\ref{fig:counterfactual_dsprite} depicts corresponding counterfactual image pairs $(Y(0),Y(1))$, confirming that the treatment $A=1$ only shifts the heart horizontally without altering overall brightness or shape.

\paragraph{Adaptive data collection.}
Logged trajectories $\{(X_t,A_t,Y_t)\}_{t=1}^T$ are generated by an $\varepsilon$-greedy contextual policy with two arms and per-arm online ridge regression, identical to the adaptive linear setting in~\S\ref{app:synthetic}. Each arm $a\in\{0,1\}$ maintains the sufficient statistics
\[
S_a = \mathrm{diag}(0,\lambda,\ldots,\lambda),\qquad b_a = 0,
\]
with $\lambda=10^{-2}$ and features $x_t^{\mathrm{aug}}=(1,X_t)\in\mathbb{R}^3$.  
At each round $t$, the arm parameters $\theta_a=S_a^{-1}b_a$ yield predictions $q_a(t)=\langle \theta_a,x_t^{\mathrm{aug}}\rangle$. The exploration rate follows
\[
\varepsilon_t=\max(\varepsilon_{\min},\,\varepsilon_0/(t+1)^p),
\qquad \varepsilon_0=0.2,\ \varepsilon_{\min}=0.05,\ p=0.99.
\]
Actions are sampled according to
\[
\pi_t(1|X_t)=
\begin{cases}
1-\tfrac{1}{2}\varepsilon_t, & q_1(t)>q_0(t),\\[3pt]
\tfrac{1}{2}\varepsilon_t, & q_1(t)<q_0(t),\\[3pt]
0.5, & q_1(t)=q_0(t),
\end{cases}
\qquad
\pi_t(0|X_t)=1-\pi_t(1|X_t).
\]
After observing $(X_t,A_t,Y_t)$, only the chosen arm is updated:
\[
S_{A_t}\leftarrow S_{A_t}+x_t^{\mathrm{aug}}(x_t^{\mathrm{aug}})^\top,
\qquad
b_{A_t}\leftarrow b_{A_t}+x_t^{\mathrm{aug}}Y_t.
\]
The sequence $\{\pi_t(1|X_t)\}$ is stored to compute the stabilized kernel test statistics.

\paragraph{Foldwise evaluation.}
To enable cross-fold variance stabilization, we use an \emph{alternating} split $(\mathcal{I}_0,\mathcal{I}_1)$ and record fold-specific propensity matrices $\Pi_{r\leftarrow r}$ computed from the parameter snapshots $\{\theta_a^{(t)}\}_{t\in\mathcal{I}_r}$.  
Each matrix encodes, for every evaluation time $t$ in a fold, the propensities $\pi_t(A_s|X_s)$ for all contexts $s$ within the same fold.

\paragraph{Evaluation protocol.}
Each experiment runs for $T=1000$ adaptive rounds and is repeated over $200$ Monte-Carlo replications. For each test, empirical type-I error is the proportion of rejections at level $0.05$ under Scenario~I, and empirical power is the proportion of rejections under Scenario~IV.  
All tests use a Gaussian RBF kernel on outcomes with bandwidth chosen by the median heuristic and $\lambda=10^{-2}$ regularization. \textsc{ADR-KTE} operates directly on flattened images $Y_t\in\mathbb{R}^{4096}$, while baseline methods (\textsc{CADR}, \textsc{AW-AIPW}) are restricted to the mean pixel intensity as scalar outcome.

\subsection{Additional results}
\label{appendix:AdditionalNumericalResults}
\paragraph{Synthetic dataset.}
To complete the presentation of our synthetic dataset experiments, this section provides the comparative results for our proposed method and the baseline algorithms under two alternative potential outcome generating functions: the linear model and the sigmoidal model, both discussed in Section \ref{app:synthetic}.

\begin{figure*}[ht]
  \centering
  \includegraphics[width=0.89\textwidth]{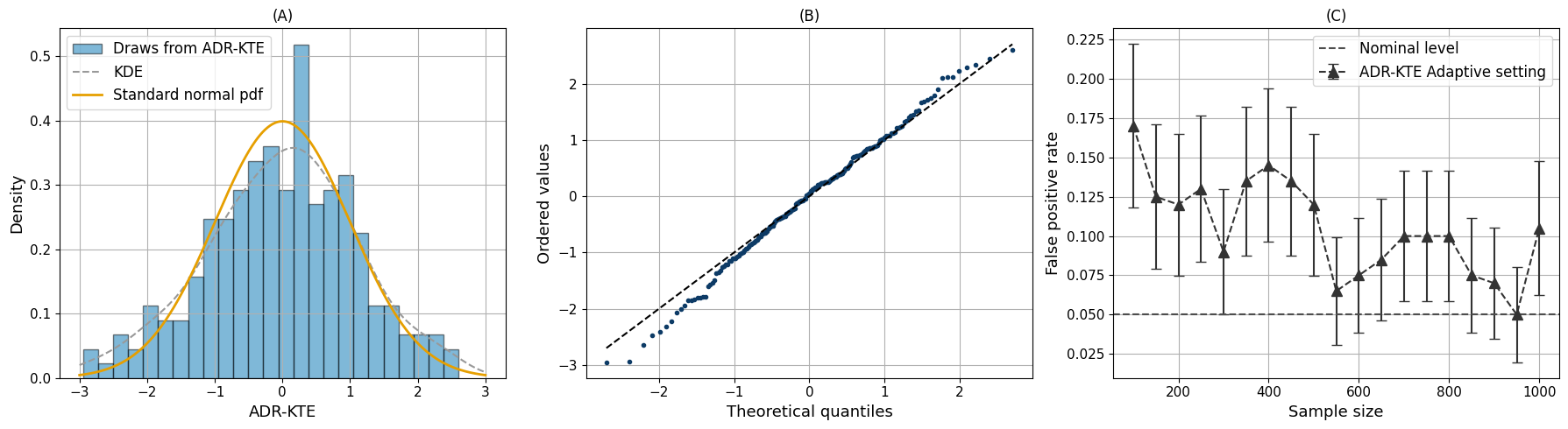}
\caption{Calibration of ADR-KTE under the null hypothesis (Scenario I) in the adaptive setting for the linear model (based on $200$ simulations). (A): Empirical histogram vs. standard normal PDF ($T=900$); (B): Normal Q-Q plot; (C): False Positive Rate across sample sizes. The results confirm approximate Gaussian asymptotics and controlled type-I error.}
\label{fig:null_adaptive_linear_synthetic}
\end{figure*}

\begin{figure*}[ht]
  \centering
\includegraphics[width=0.89\linewidth]{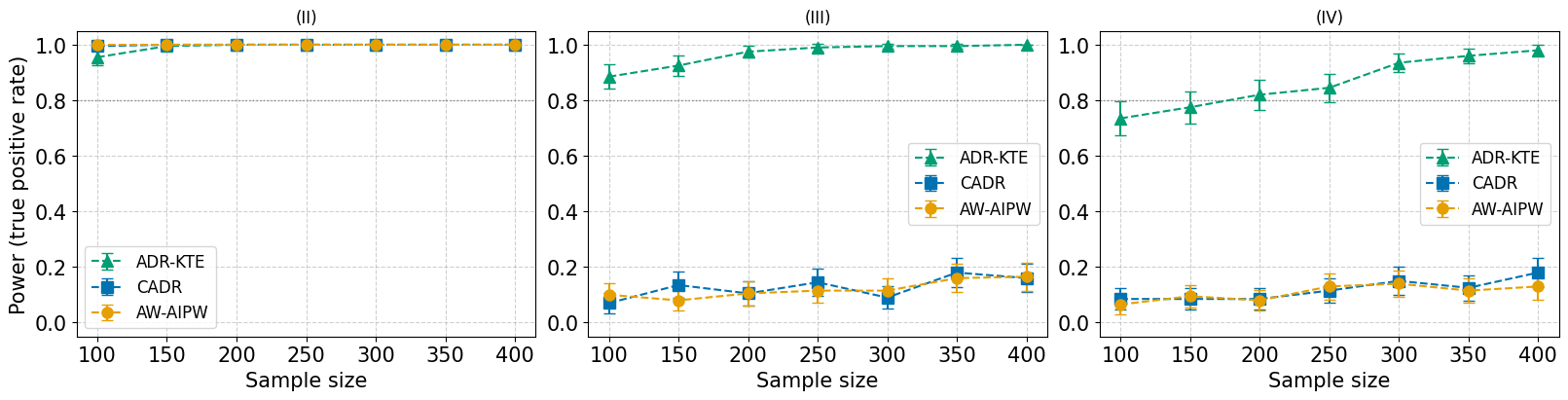}
\caption{Power comparison (true positive rates) for the linear model across Scenarios II–IV, based on $200$ simulations. Mean-focused baselines (CADR/AW-AIPW) achieve matching power on Scenario II (mean shift). ADR-KTE demonstrates markedly higher power in detecting higher-moment shifts (Scenarios III–IV).}
\label{fig:adaptive_power_linear_synthetic}
\end{figure*}

\begin{figure*}[ht]
  \centering
  \includegraphics[width=0.89\textwidth]{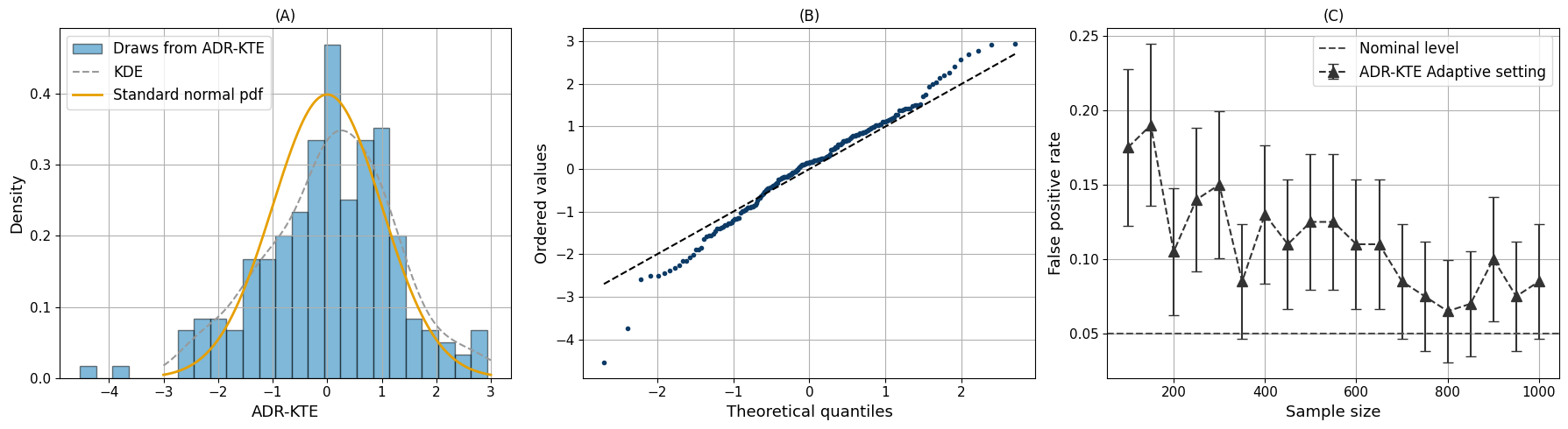}
\caption{Demonstration of the Calibration of ADR-KTE in the adaptive setting for the sigmoidal model under the null hypothesis (Scenario I), based on $200$ replications. (A): Histogram of test statistics compared to the standard normal PDF (shown for $T=850$); (B): Normal Q-Q plot; (C): type-I error (False Positive Rate) evolution across sample sizes.}
\label{fig:null_adaptive_sigmoid_synthetic}
\end{figure*}

\begin{figure*}[ht]
  \centering
\includegraphics[width=0.89\linewidth]{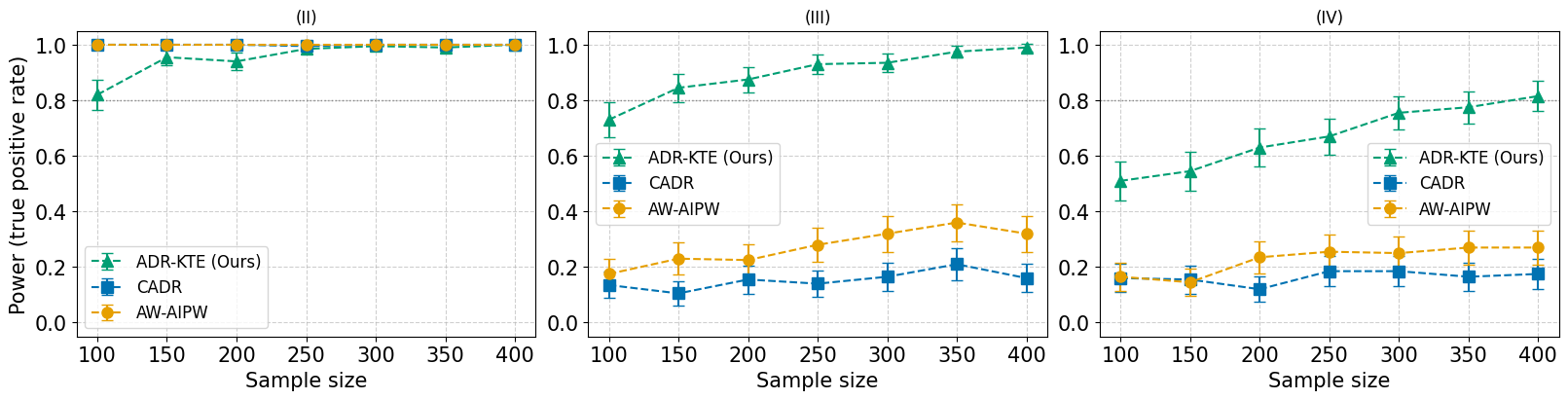}
\caption{Comparative Power results (true positive rates) for the sigmoidal model across Scenarios II–IV, using $200$ Monte-Carlo runs. Baselines focused on mean effects (CADR/AW-AIPW) achieve matching performance for the mean shift in Scenario II. In contrast, ADR-KTE displays a significantly greater ability to detect distributional differences characterized by higher-moment shifts (Scenarios III–IV).}
\label{fig:adaptive_power_sigmoid_synthetic}
\end{figure*}

\begin{itemize}
    \item \textbf{Linear model results:} The calibration of our proposed method, ADR-KTE, in the linear case (Scenario I) is demonstrated in Figure \ref{fig:null_adaptive_linear_synthetic}. The collected metrics—including the empirical histogram, Q-Q plot, and false positive rate across varying data sizes—collectively confirm that our method is well-calibrated.

Figure \ref{fig:adaptive_power_linear_synthetic} provides the comparison of ADR-KTE with the baselines CADR and AW-AIPW across Scenarios II-IV. Consistent with our preceding findings, the baselines achieve matching performance in Scenario II (mean shift) and even show slightly better results in the small data size regime. Crucially, however, our method significantly outperforms the baselines in scenarios characterized by purely distributional changes with an identical mean (Scenarios III and IV).

\item \textbf{Sigmoidal model results:} The findings for the sigmoidal case similarly mirror these results. The calibration of ADR-KTE in Scenario I is shown in Figure \ref{fig:null_adaptive_sigmoid_synthetic}, while the comparative power results across Scenarios II-IV are displayed in Figure \ref{fig:adaptive_power_sigmoid_synthetic}. In both model structures, our method maintains its superior power in detecting distributional differences where mean-based methods fail.
\end{itemize}

\paragraph{IHDP dataset:} We now present the results from the numerical simulations conducted on the IHDP dataset, focusing on the method's performance across varying sample sizes.

\begin{figure*}[ht]
  \centering
  \includegraphics[width=0.89\textwidth]{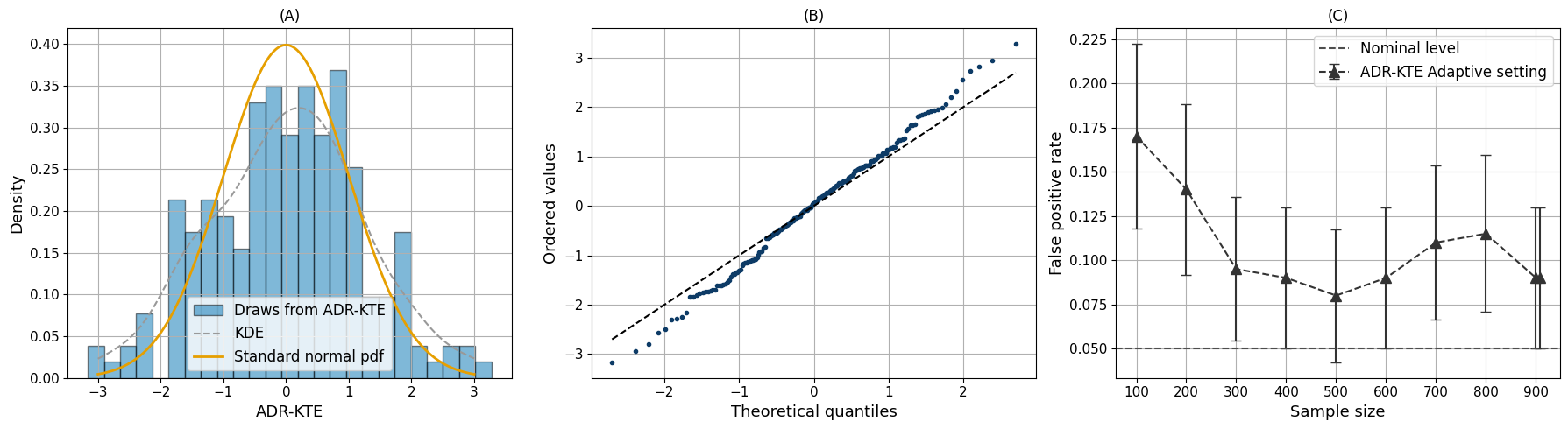}
\caption{Assessment of the Calibration of ADR-KTE under the null hypothesis (Scenario I) in the adaptive setting, using the IHDP dataset ($200$ replications). (A): Distribution of test statistics (histogram versus standard normal PDF, shown for the full sample size $T=908$); (B): Normal Q-Q plot; (C): type-I error (False Positive Rate) control across varying sample sizes.}
\label{fig:null_adaptive_ihdp}
\end{figure*}

Figure \ref{fig:null_adaptive_ihdp} illustrates the calibration of our proposed method under the null hypothesis (Scenario I), based on $200$ Monte-Carlo runs. This figure presents the histogram of test statistics, the Q-Q plot, and the type-I error across varying sample sizes.

\begin{figure*}[ht]
  \centering
\includegraphics[width=0.89\linewidth]{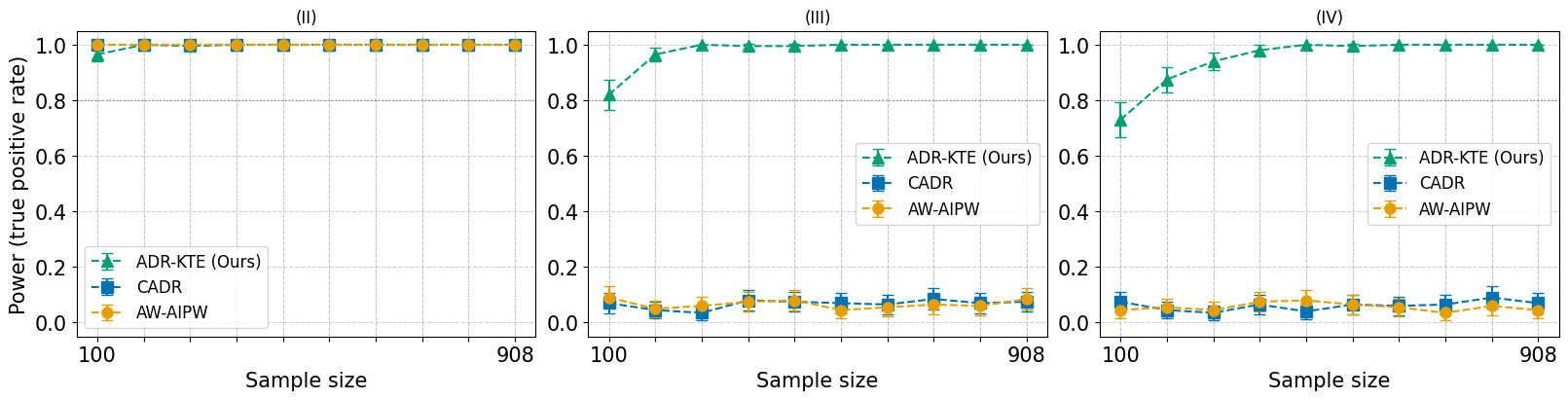}
\caption{Comparative Power Analysis (true positive rates) for the IHDP dataset across Scenarios II–IV, based on $200$ Monte-Carlo runs. The mean-focused baselines (CADR/AW-AIPW) show matching detection capability for the pure mean shift in Scenario II. Conversely, ADR-KTE exhibits a substantially improved power profile for identifying distributional disparities stemming from higher-moment changes (Scenarios III–IV).}
\label{fig:adaptive_power_ihdp}
\end{figure*}

The power of our method in comparison with the baselines for Scenarios II-IV is demonstrated across varying data sizes in Figure \ref{fig:adaptive_power_ihdp}. These results show that, in particular, our method exhibits a significant advantage in power for detecting distributional effects, in contrast to the mean-focused baselines.

\subsection{Comparison and Discussion with Non-adaptive Kernel Baselines}
\label{app:iid_comparison}

To rigorously assess the impact of data adaptivity on standard kernel methods, we benchmarked our \textsc{ADR-KTE} test against two prominent estimators designed for i.i.d.\ data: \textsc{DR-xKTE} \citep{martinez2023} and the standard \textsc{KTE} \citep{muandet2021counterfactual}. We evaluate these across $T \in \{100, 150, \dots, 500\}$ to demonstrate the stability of our approach under varying sample sizes.

\paragraph{Data Generating Process.} 
For both i.i.d.\ and adaptive settings, we utilize a sigmoidal baseline function $h(x) = \operatorname{sgn}(x-0.5) \log(|16x-8|+1)$ to generate the outcome base. The potential outcomes are then defined as $Y_t(0) = h(X_t^\top \beta) + \varepsilon_t$ and $Y_t(1) = h(X_t^\top \beta) + \Delta_t + \varepsilon_t$, where $\beta = [0.1, 0.2, 0.3, 0.4, 0.5]^\top$ and $\varepsilon_t \sim \mathcal{N}(0, 0.5)$. In the i.i.d.\ case, actions are sampled from a fixed logistic policy $\pi(1|x) = \operatorname{logit}^{-1}(x^\top \beta)$. In the adaptive case, we employ the $\varepsilon$-greedy contextual bandit described in Section \ref{sec:SyntheticDataNumExperiment}, which updates its arm parameters and exploration rate $\varepsilon_t$ at every time step $t$.

\paragraph{Case 1: Verification under i.i.d.\ Sampling}
We first verify that all estimators are correctly calibrated under a standard i.i.d.\ sampling scheme. As shown in Table~\ref{tab:iid_scheme_results}, all procedures exhibit nominal type-I error under the null (Scenario I) and full power under the alternative (Scenario II). This confirms that in the absence of adaptivity, our proposed \textsc{ADR-KTE} maintains the standard effectiveness of i.i.d.\ methods.
\begin{table}[h]
\centering
\caption{Rejection rates (mean $\pm$ se) under an i.i.d.\ sampling scheme for Scenario I (Null) and Scenario II (Alternative). All methods are correctly calibrated and achieve full power when data collection is not adaptive.}
\label{tab:iid_scheme_results}
\begin{adjustbox}{width=0.9\textwidth}
\begin{tabular}{l l l l l l l l l l}
\toprule
\textbf{Scenario I (Null)} & $100$ & $150$ & $200$ & $250$ & $300$ & $350$ & $400$ & $450$ & $500$ \\
\midrule
\textsc{DR-xKTE} & $0.085$ & $0.105$ & $0.050$ & $0.045$ & $0.075$ & $0.065$ & $0.055$ & $0.075$ & $0.055$ \\
\textsc{KTE} & $0.050$ & $0.010$ & $0.025$ & $0.055$ & $0.055$ & $0.070$ & $0.040$ & $0.025$ & $0.040$ \\
\textsc{ADR-KTE} & $0.085$ & $0.040$ & $0.090$ & $0.085$ & $0.050$ & $0.080$ & $0.085$ & $0.070$ & $0.065$ \\
\midrule
\textbf{Scenario II (Alt)} & $100$ & $150$ & $200$ & $250$ & $300$ & $350$ & $400$ & $450$ & $500$ \\
\midrule
\textsc{DR-xKTE} & $1.000$ & $1.000$ & $1.000$ & $1.000$ & $1.000$ & $1.000$ & $1.000$ & $1.000$ & $1.000$ \\
\textsc{KTE} & $1.000$ & $1.000$ & $1.000$ & $1.000$ & $1.000$ & $1.000$ & $1.000$ & $1.000$ & $1.000$ \\
\textsc{ADR-KTE} & $1.000$ & $1.000$ & $1.000$ & $1.000$ & $1.000$ & $1.000$ & $1.000$ & $1.000$ & $1.000$ \\
\bottomrule
\end{tabular}
\end{adjustbox}
\end{table}

\paragraph{Case 2: Evaluation under Stable Adaptive Data Collection.}
We next evaluate the methods under the $\varepsilon$-greedy adaptive policy. In contrast to highly anisotropic exploration rules, this policy maintains a uniform exploration floor, so the empirical action-feature covariance is naturally viewed as operating in a stable adaptive regime, and in our setting it is reasonable to regard it as satisfying the stronger full-matrix stability condition \citep{laiweiaos1982}. Recent theory shows that, under such stability conditions, estimators that are efficient in the i.i.d.\ setting remain asymptotically normal and efficient under adaptive sampling for scalar pathwise differentiable targets \citep{shen2026}. Although that result is stated for scalar targets, the underlying mechanism is the stabilization of the predictable quadratic variation, which suggests that an analogous phenomenon should also hold for Hilbert-valued and RKHS-based scores.

Table~\ref{tab:adaptive_scheme_results} is broadly consistent with this picture, while also showing that finite-sample behavior remains method-dependent. \textsc{DR-xKTE} is clearly anti-conservative under the null at small and moderate horizons, so its large rejection rates under the alternative are not directly interpretable as reliable power. \textsc{KTE} stays roughly calibrated, but suffers a substantial loss of power throughout. In contrast, \textsc{ADR-KTE} delivers by far the strongest overall performance: it attains high power across all horizons, and its null rejection rate decreases steadily toward the nominal level as $T$ grows. Overall, these results support the view that our $\varepsilon$-greedy design lies in a stable adaptive regime, and that explicit variance control remains important in practice for RKHS-based inference, even when the underlying adaptive scheme is asymptotically well-behaved.

\begin{table}[h]
\centering
\caption{Rejection rates under the $\varepsilon$-greedy adaptive policy. Because this policy maintains persistent exploration, the resulting adaptive design is naturally viewed as stable, plausibly in the stronger full-matrix sense. The table shows that finite-sample performance nevertheless depends strongly on the test statistic: \textsc{DR-xKTE} is anti-conservative under the null, \textsc{KTE} is roughly calibrated but markedly underpowered, and \textsc{ADR-KTE} provides the best calibration-power tradeoff, with null rejection rates moving toward the nominal level as $T$ increases while power remains high.}
\label{tab:adaptive_scheme_results}
\begin{adjustbox}{width=0.9\textwidth}
\begin{tabular}{l l l l l l l l l l}
\toprule
\textbf{Scenario I (Null)} & $100$ & $150$ & $200$ & $250$ & $300$ & $350$ & $400$ & $450$ & $500$ \\
\midrule
\textsc{DR-xKTE} & $0.165$ & $0.150$ & $0.125$ & $0.100$ & $0.080$ & $0.090$ & $0.105$ & $0.095$ & $0.075$ \\
\textsc{KTE} & $0.065$ & $0.055$ & $0.055$ & $0.035$ & $0.040$ & $0.045$ & $0.060$ & $0.035$ & $0.035$ \\
\textsc{ADR-KTE} & $0.140$ & $0.125$ & $0.125$ & $0.120$ & $0.145$ & $0.115$ & $0.065$ & $0.095$ & $0.060$ \\
\midrule
\textbf{Scenario II (Alt)} & $100$ & $150$ & $200$ & $250$ & $300$ & $350$ & $400$ & $450$ & $500$ \\
\midrule
\textsc{DR-xKTE} & $0.960$ & $1.000$ & $0.985$ & $0.995$ & $1.000$ & $1.000$ & $1.000$ & $1.000$ & $1.000$ \\
\textsc{KTE} & $0.260$ & $0.200$ & $0.220$ & $0.180$ & $0.155$ & $0.225$ & $0.145$ & $0.190$ & $0.245$ \\
\textsc{ADR-KTE} & $0.885$ & $0.945$ & $0.950$ & $0.955$ & $0.990$ & $0.995$ & $1.000$ & $1.000$ & $1.000$ \\
\bottomrule
\end{tabular}
\end{adjustbox}
\end{table}

\newpage

\subsection{Computation infrastructure}

We ran our experiments on local CPUs of desktops and on a GPU-enabled node (in a remote server) with the following specifications:

\begin{itemize}[label=\textbullet]
\item \textbf{Operating System:} Linux (kernel version 6.8.0-55-generic)
\item \textbf{GPU:} NVIDIA RTX A4500
\begin{itemize}
\item Driver Version: 560.35.05
\item CUDA Version: 12.6
\item Memory: 20 GB GDDR6
\end{itemize}
\end{itemize}

\end{document}